\documentclass[11pt]{article}

\usepackage[a4paper,margin=1in]{geometry}
\usepackage[T1]{fontenc}
\usepackage[utf8]{inputenc} % OK for pdfLaTeX/Overleaf
\usepackage{lmodern}
\usepackage{microtype}

\usepackage{amsmath,amssymb}
\usepackage{graphicx}
\usepackage[table]{xcolor}
\usepackage{adjustbox}

\definecolor{grpGenerative}{HTML}{d62728}
\definecolor{grpToolAug}{HTML}{ff7f0e}
\definecolor{grpAgentic}{HTML}{2ca02c}
\definecolor{grpWfNL}{HTML}{9467bd}
\definecolor{grpWfCentric}{HTML}{1f77b4}

\definecolor{thDeterminism}{HTML}{1f77b4}
\definecolor{thFlexibility}{HTML}{ff7f0e}
\definecolor{thCrossCut}{HTML}{2ca02c}
\definecolor{thOutScope}{HTML}{7f7f7f}

\usepackage{booktabs}
\usepackage{array}
\usepackage{tabularx}
\usepackage{multirow}
\usepackage{chngcntr}
\counterwithout{table}{section}
\counterwithout{figure}{section}
\usepackage{caption}
\usepackage{subcaption}

\usepackage{enumitem}
\usepackage{csquotes}
\usepackage{float}
\usepackage{placeins} % provides \FloatBarrier to keep floats from drifting past a point
\usepackage{xspace}

\usepackage{hyperref}
\hypersetup{
  colorlinks=true,
  linkcolor=blue,
  citecolor=blue,
  urlcolor=blue,
  pdftitle={Talk Freely, Execute Strictly: Schema-Gated Agentic AI for Flexible and Reproducible Scientific Workflows},
  pdfauthor={Joel Strickland; Arjun Vijeta; Chris Moores; Oliwia Bodek; Bogdan Nenchev; Thomas Whitehead; Charles Phillips; Karl Tassenberg; Gareth Conduit; Ben Pellegrini}
}

\usepackage[numbers,sort&compress]{natbib}

\newcommand{\tm}{\texttrademark}
\newcommand{\cf}{CF\xspace}
\newcommand{\ed}{ED\xspace}

\setlist[itemize]{noitemsep, topsep=4pt}
\setlist[enumerate]{noitemsep, topsep=4pt}

\title{\textbf{Talk Freely, Execute Strictly: Schema-Gated Agentic AI for Flexible and Reproducible Scientific Workflows}}

\author{
  Joel Strickland$^{1}$\thanks{Corresponding author: \href{mailto:joel@intellegens.com}{joel@intellegens.com}}
  \and Arjun Vijeta$^{1}$
  \and Chris Moores$^{2}$
  \and Oliwia Bodek$^{2}$
  \and Bogdan Nenchev$^{1}$
  \and Thomas Whitehead$^{1}$
  \and Charles Phillips$^{1}$
  \and Karl Tassenberg$^{1}$
  \and Gareth Conduit$^{1,3}$
  \and Ben Pellegrini$^{1}$
}

\date{
  \small
  $^{1}$ Intellegens, The Studio, Chesterton Mill, Cambridge, CB4 3NP, UK\\
  $^{2}$ Independent Researcher\\
  $^{3}$ Cavendish Laboratory, J.J. Thomson Avenue, Cambridge, UK
}

\begin{document}
\makeatletter
% Encourage floats (tables/figures) to appear closer to where they are mentioned,
% rather than collecting at the end of the document.
\setcounter{topnumber}{4}
\setcounter{bottomnumber}{2}
\setcounter{totalnumber}{6}
\renewcommand{\topfraction}{0.9}
\renewcommand{\bottomfraction}{0.8}
\renewcommand{\textfraction}{0.07}
\renewcommand{\floatpagefraction}{0.8}
\makeatother
\maketitle

% ---------- Abstract ----------
\begin{abstract}
\sloppy
Large language models (LLMs) can now translate a researcher's plain-language goal into executable computation, yet scientific workflows demand determinism, provenance, and governance that are difficult to guarantee when an LLM decides what runs. Semi-structured interviews with eighteen experts across ten industrial R\&D stakeholders surface two competing requirements---deterministic, constrained execution and conversational flexibility without workflow rigidity---together with boundary properties (human-in-the-loop control and transparency/provenance) that any resolution must satisfy. We propose \emph{schema-gated orchestration} as the resolving principle: the schema becomes a mandatory execution boundary at the composed-workflow level, so that nothing runs unless the complete action---including cross-step dependencies---validates against a machine-checkable specification.

We operationalise the two requirements as execution determinism (\ed) and conversational flexibility (\cf), and use these axes to review 20 systems spanning five architectural groups along a validation-scope spectrum. Scores are assigned via a multi-model protocol---15 independent sessions across three LLM families---yielding substantial-to-near-perfect inter-model agreement (Krippendorff's $\alpha = 0.80$ for \ed{} and $\alpha = 0.98$ for \cf), demonstrating that multi-model LLM scoring can serve as a reusable alternative to human expert panels for architectural assessment.

The resulting landscape reveals an empirical Pareto front---no reviewed system achieves both high flexibility and high determinism---but a convergence zone emerges between the generative and workflow-centric extremes. We argue that a schema-gated architecture, separating conversational from execution authority, is positioned to decouple this trade-off, and distil three operational principles---clarification-before-execution, constrained plan--act orchestration, and tool-to-workflow-level gating---to guide adoption.
\end{abstract}
\section{Introduction}
Modern scientific discovery depends on computational workflows that chain diverse tools---from data preparation through modelling and analysis \cite{Strickland2021,Strickland2020,Strickland2025,CohenBoulakia2017,Hope2023,Liu2025}. Progress is combinatorial: researchers swap simulation engines, test alternative featurizations, and evaluate competing models as evidence accumulates \cite{Huber2021,Ganose2025,Elton2018,Patel2022,Lamprecht2021,Singh2021,Gustafsson2025,Strickland2024}. Yet the supporting software ecosystems remain fragmented \cite{Farshidi2023,McInnes2025}---tools are separated by format mismatches \cite{OliveiraStewart2006,Alam2025a}, inconsistent parameter semantics \cite{Lee2018,Wang2023a}, and incompatible execution models \cite{Pimentel2008}. Many analyses are therefore assembled through ad hoc pipelines with poorly documented parameter choices, software versions, and execution environments \cite{Sandve2013,StoddenMiguez2014,Jimenez2017}, undermining reproducibility and auditability \cite{Freire2008,Missier2013,Moreau2011,Boettiger2015,Gruning2018,Hannay2009,Hettrick2014,Krafczyk2021,Ziemann2023,deOliveiraAndrade2025}.

Large language models (LLMs) offer a conversational alternative: users describe goals in natural language while the system selects and executes tools \cite{YildizPeterka2024,AlamRoy2025,Jiang2025,Yao2022,Wang2024a,Bran2024,Shin2025}. Yet prompt-driven code generation introduces reproducibility and provenance risks: ad hoc implementations can vary across runs \cite{MorishigeKoshihara2025,Hosseini2025}, identical prompts can produce different algorithms or parameter defaults \cite{Baltes2025,BrucksToubia2025,Kervadec2023}, and the resulting lack of traceability undermines trust \cite{Ji2023}. Governance risks---prompt injection, data leakage, untracked supply-chain dependencies---further complicate deployment unless validation and auditability are enforced by design \cite{Jia2025,Hines2024,Hu2025,Das2025}.

Traditional workflow management systems such as Galaxy \cite{Goecks2010}, Snakemake \cite{KosterRahmann2012}, and Nextflow \cite{DiTommaso2017} address many of these concerns by requiring explicit workflow specifications. They provide strong determinism, provenance, and scalability guarantees, but impose substantial interaction costs and limited support for conversational or exploratory use \cite{Molder2021,Kanwal2017}. Contemporary approaches therefore split into two tendencies: generative, code-centric systems that maximize flexibility at the expense of reproducibility, and workflow-centric systems that ensure reproducibility at the expense of conversational use.

We frame this tension through the concept of \textit{execution authority}---where responsibility for concrete executable behavior resides and what guarantees must hold before computation runs (\S\ref{sec:exec-authority}). Semi-structured interviews with practitioners from ten industrial R\&D stakeholders (\S\ref{sec:interviews}) surface two competing requirements---execution determinism (Req\,A) and conversational flexibility (Req\,B)---which we use to map 20 representative systems in an \ed/\cf design space (\S\ref{sec:ordinal-scoring}). The review reveals a convergence zone; we term the underlying principle \emph{schema-gated orchestration} and instantiate it in a reference architecture (\S\ref{sec:architecture}).

\paragraph{Relationship to prior work.}
The closest related framework is the ``(R)evolution'' roadmap of Shin et~al.~\cite{Shin2025}, which charts a path from current workflow systems to fully autonomous laboratories along two dimensions---intelligence (static to intelligent) and composition (single to swarm)---and proposes a federated blueprint for autonomous science. Our work is complementary but differs in three respects: (i)~we ground the design space in practitioner requirements (18~experts, 10~stakeholders) rather than a technology-forward vision, surfacing the dominance of integration and determinism concerns over autonomy aspirations; (ii)~our \ed/\cf axes capture a specific trade-off---the coupling of conversational and execution authority---that the intelligence/composition dimensions do not address (Shin et~al.\ flag GenAI non-determinism vs.\ reproducibility as an open challenge but do not operationalise it); and (iii)~we identify schema-gated orchestration as the resolving mechanism, a pattern absent from their blueprint. Where their framework asks \emph{how intelligent and how composed} a system should be, ours asks \emph{where execution authority resides and what must validate before computation runs}. The intelligence/composition matrix charts the long-horizon evolution of workflow capabilities; \ed/\cf and schema-gating address the near-term question of how to make LLM-driven systems trustworthy for scientific use.

\paragraph{Contributions.}
\begin{enumerate}[nosep]
  \item A thematic analysis of practitioner needs from ten industrial R\&D stakeholders, surfacing a tension between execution determinism (Req\,A) and conversational flexibility (Req\,B) together with boundary properties any resolution must satisfy (\S\ref{sec:interviews}).
  \item A five-group taxonomy of 20 representative systems along a validation-scope spectrum, scored on ordinal \ed/\cf axes and visualised in an architectural design space that reveals an empirical Pareto front (\S\ref{sec:ordinal-scoring}).
  \item A reusable multi-model scoring protocol---15 independent sessions across three LLM families with cross-model agreement analysis---demonstrating that LLM-based architectural assessment can achieve substantial-to-near-perfect inter-rater reliability (\S\ref{sec:ordinal-scoring}; Appendix~\ref{app:scoring-protocol}).
  \item A synthesis of \emph{schema-gated execution} as an architectural principle separating conversational from execution authority via schema-validated actions, together with three operational principles: clarification-before-execution, constrained plan--act orchestration, and tool-to-workflow-level gating (\S\ref{sec:architecture}).
  \item A reference schema-gated architecture with validated execution and end-to-end provenance capture, presented as an architectural proposal whose empirical validation is identified as the most pressing near-term priority (\S\ref{sec:ref-arch}; \S\ref{sec:discussion}; \S\ref{sec:future-work}, item~4).
\end{enumerate}

% ---------- 2. User Research and Requirements Elicitation ----------
\section{User Research and Requirements Elicitation}
\label{sec:interviews}
To understand what industrial R\&D practitioners require from conversational AI workflow systems, we conducted semi-structured interviews (typically two sessions per stakeholder, each $\sim$30~minutes; 20~sessions total) with 18~experts across ten stakeholders recruited through Intellegens' professional network, spanning specialty chemicals, food science, materials and packaging, semiconductor processing, construction, and manufacturing (Table~\ref{tab:participant-demographics}). Sessions covered workflow challenges, automation opportunities, and expectations for agentic AI in R\&D. Interviews were recorded with consent and transcribed. Themes were coded from raw transcripts using regex-based pattern matching against an \emph{a~priori} codebook, with interviewer and internal-team utterances excluded. Because all stakeholders were recruited through Intellegens' professional network, the sample is one of convenience and may over-represent organisations already investing in ML-assisted workflows; this limits generalisability and is discussed further in Appendix~\ref{app:interview-methods}. Full recruitment details, confidentiality protocols, interview structure, and the coding protocol are provided in Appendix~\ref{app:interview-methods}.

\subsection{Thematic landscape}
\label{sec:thematic-landscape}

Responses were coded using systematic content coding informed by Braun and Clarke~\cite{BraunClarke2006}, yielding 17 themes from 1{,}135 codings across 2{,}468 speaker-turn paragraphs (18 experts, 10 stakeholders). To ensure balanced extraction, each theme is defined by a fixed set of 20~regex patterns---the minimum at which the narrowest themes (e.g., Progressive Autonomy) achieved adequate recall. Fixing the count prevents lexically rich themes from dominating through vocabulary breadth, at the cost of potentially limiting precision for sparse themes. The full codebook is provided in Appendix~\ref{app:interview-methods}; a three-part sensitivity analysis (Appendix~\ref{app:sensitivity})---leave-one-out stability, cross-meeting convergent validity, and bootstrap confidence intervals---confirmed robust theme rankings, with spot-check precision of 86--93\%.

The 17 themes cluster into four groups based on their architectural implications (Table~\ref{tab:thematic-groups}).

\begin{table}[t]
  \centering
  \small
  \caption{Themes from practitioner interviews grouped by architectural implication. \emph{Mentions} is the total count across all experts; \emph{Breadth} is the number of stakeholders (out of 10) in which the theme appeared. Group headers show total mentions and per-theme average ($\bar{x}$) to allow comparison across groups of different size. Row colour encodes thematic group: blue~= determinism pole, orange~= flexibility pole, green~= cross-cutting, grey~= out of scope (matching Figure~\ref{fig:figure1}).}
  \label{tab:thematic-groups}
  \setlength{\tabcolsep}{5pt}
  \renewcommand{\arraystretch}{1.1}
  \begin{tabularx}{\linewidth}{@{}l X r r@{}}
    \toprule
    \textbf{Group} & \textbf{Theme} & \textbf{Mentions} & \textbf{Breadth} \\
    \midrule
    \rowcolor{thDeterminism!10}
      & System Integration          & 191 & 10/10 \\
    \rowcolor{thDeterminism!10}
      & Workflow \& Task Automation  & 112 & 10/10 \\
    \rowcolor{thDeterminism!10}
      & Data Cleaning \& Preparation & 70 & 10/10 \\
    \rowcolor{thDeterminism!10}
      & Security \& IP Protection    & 67 & 9/10 \\
    \rowcolor{thDeterminism!10}
      & Domain / Scientific Expertise & 62 & 9/10 \\
    \rowcolor{thDeterminism!10}
    \multirow{-6}{*}{\parbox{3.8cm}{\raggedright\textbf{Determinism pole}\\(527 total; $\bar{x}=88$)}}
      & Safety \& Ethical Boundaries  & 25 & 9/10 \\
    \addlinespace[12pt]
    \rowcolor{thFlexibility!10}
      & Data Search \& Retrieval       & 108 & 10/10 \\
    \rowcolor{thFlexibility!10}
      & Natural Language Interface     & 86 & 10/10 \\
    \rowcolor{thFlexibility!10}
      & Visualisation \& Reporting     & 65 & 9/10 \\
    \rowcolor{thFlexibility!10}
    \multirow{-4}{*}{\parbox{3.8cm}{\raggedright\textbf{Flexibility pole}\\(300 total; $\bar{x}=75$)}}
      & Multi-Agent / Agent-to-Agent   & 41 & 9/10 \\
    \addlinespace[12pt]
    \rowcolor{thCrossCut!10}
      & Explainability \& Transparency & 83 & 9/10 \\
    \rowcolor{thCrossCut!10}
      & Human Oversight \& Control     & 44 & 9/10 \\
    \rowcolor{thCrossCut!10}
    \multirow{-3}{*}{\parbox{3.8cm}{\raggedright\textbf{Cross-cutting / boundary properties}\\(146 total; $\bar{x}=49$)}}
      & Progressive Autonomy           & 19 & 5/10 \\
    \addlinespace[12pt]
    \rowcolor{thOutScope!10}
      & Cost \& Efficiency                  & 58 & 9/10 \\
    \rowcolor{thOutScope!10}
      & Adoption \& Change Management       & 51 & 10/10 \\
    \rowcolor{thOutScope!10}
      & Deployment Model Preferences        & 43 & 10/10 \\
    \rowcolor{thOutScope!10}
    \multirow{-4}{*}{\parbox{3.8cm}{\raggedright\textbf{Out of arch.\ scope}\\(162 total; $\bar{x}=41$)}}
      & Data Aggregation \& Anonymisation   & 10 & 4/10 \\
    \bottomrule
  \end{tabularx}
\end{table}

The determinism-pole themes demand stable, reproducible, validated execution; the flexibility-pole themes demand conversational, natural-language-driven (NL) interaction for exploration and rapid iteration. The three cross-cutting themes describe properties expected at the \emph{transition} from exploration to execution, rather than aligning with either pole. Out-of-scope themes concern deployment and business considerations, discussed briefly in \S\ref{sec:discussion}.

\begin{figure}[t]
  \centering
  \includegraphics[width=0.85\columnwidth]{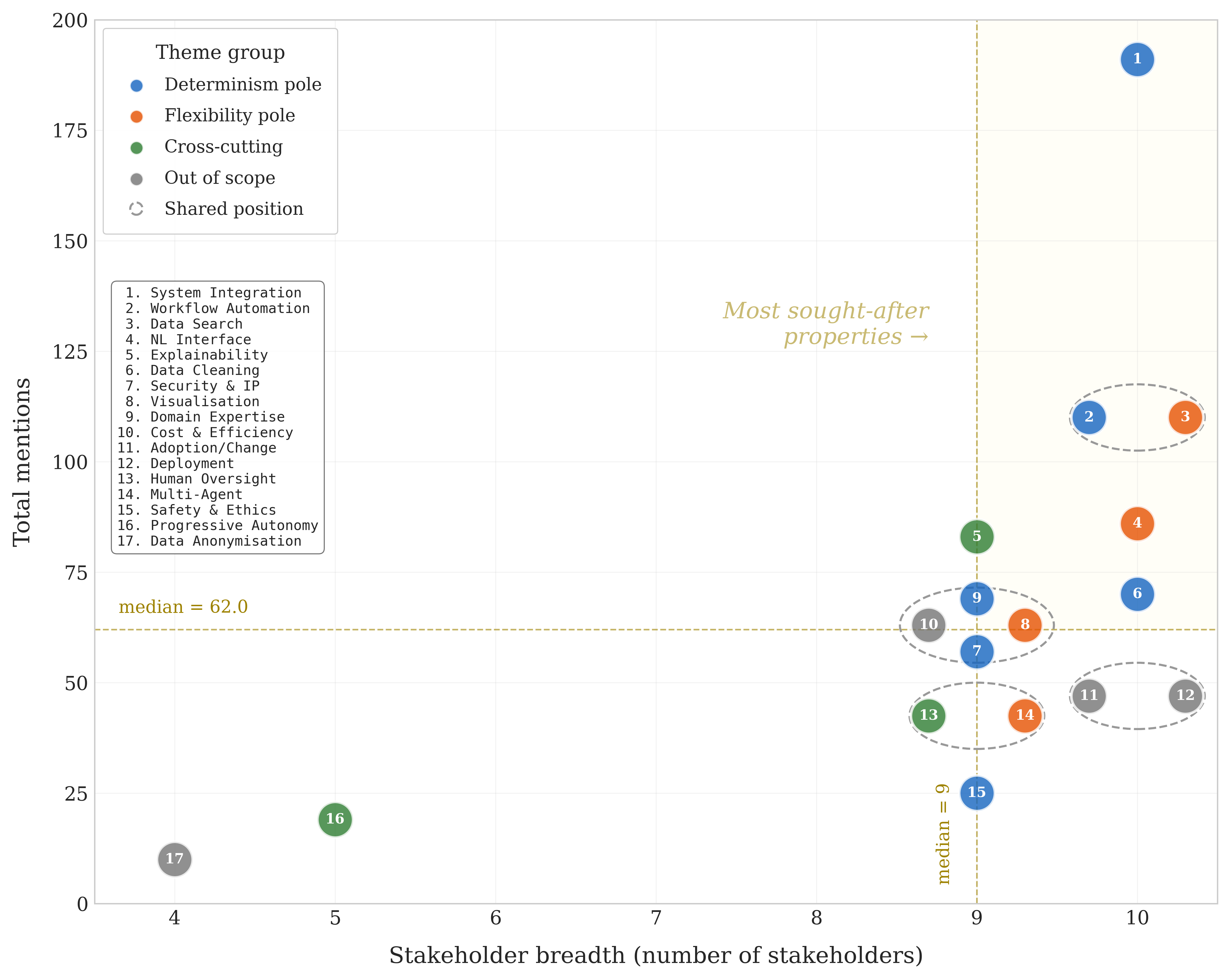}
  \caption{Thematic landscape from practitioner interviews. Each numbered circle represents one of the 17~themes; the key maps numbers to theme names. Position shows stakeholder breadth ($x$-axis: number of stakeholders mentioning the theme, out of~10) and total mention count ($y$-axis). Colour encodes theme group: blue~= determinism pole, orange~= flexibility pole, green~= cross-cutting / boundary properties, grey~= out of architectural scope. Dashed lines mark the median stakeholder breadth and median total mentions; the shaded quadrant highlights themes above both medians. Dotted ellipses enclose themes that share approximately the same grid position.}
  \label{fig:figure1}
\end{figure}

Figure~\ref{fig:figure1} visualises this landscape. \emph{System Integration} is the most prominent theme (191 mentions, 10/10 stakeholders), followed by \emph{Workflow \& Task Automation} (112, 10/10) and \emph{Data Search \& Retrieval} (108, 10/10). Flexibility-pole themes achieve near-universal breadth but lower mention intensity. Cross-cutting themes score highly on both axes, reinforcing their boundary-property status.

\subsection{Two architectural requirements}
\label{sec:requirements}
The two poles define a tension that recurred across all 10 stakeholders. Their per-theme averages ($\bar{x}=88$ for determinism, $\bar{x}=75$ for flexibility; Table~\ref{tab:thematic-groups}) substantially exceed those of the cross-cutting ($\bar{x}=49$) and out-of-scope ($\bar{x}=41$) groups, confirming that these two concerns dominated practitioner discourse even after normalising for group size. We state them as architectural requirements because they constrain the design space independent of any particular implementation.

\paragraph{Req\,A.\ Execution determinism.}
Derived from the six determinism-pole themes (527 total mentions; lead theme shared by all 10 stakeholders; Table~\ref{tab:thematic-groups}, Figure~\ref{fig:figure1}). Practitioners valued NL interaction for expressing intent but expected computations contributing to the scientific record to be stable, repeatable, and grounded in well-defined operations. Several stakeholders described a progressive trust model: initial results are manually validated, and oversight is relaxed only after the system demonstrates consistent behaviour. One chemistry team distinguished explicitly between informal exploration (``playing with a chatbot'') and organisational science, noting that the latter demands reproducible, transparent pipelines.

\paragraph{Req\,B.\ Conversational flexibility.}
Derived from the four flexibility-pole themes (300 total mentions; 9--10/10 stakeholder breadth; Table~\ref{tab:thematic-groups}, Figure~\ref{fig:figure1}). Practitioners consistently demanded rapid iteration---trying alternatives, swapping tools, and refining analyses without rewriting rigid pipelines. One materials-science stakeholder noted that roughly three-quarters of formulation work involves recombining known components with small variations, requiring quick substitution rather than pipeline redesign. Researchers wanted to describe objectives conversationally and have the system translate intent into executable steps.

\paragraph{Boundary properties.}
The three cross-cutting themes---explainability and transparency (83 mentions, 9/10 stakeholders), human-in-the-loop control (44, 9/10), and progressive autonomy (19, 5/10; Table~\ref{tab:thematic-groups})---are neither a third pole nor optional extras. A system can be fully deterministic yet offer no human approval step, or fully conversational yet completely opaque. When execution is mediated by explicit, structured invocation objects, approval gates become trivial (the invocation is an inspectable artefact) and provenance is captured by construction (every run is a validated, versioned record). These are therefore \emph{boundary properties}---design constraints that any resolution of the Req\,A/Req\,B tension must satisfy.

\paragraph{Operationalisation.}
\label{sec:req-to-axes}
We operationalise Req\,A as \emph{execution determinism} (\ed)---the strength of mandatory pre-execution constraints and the availability of replayable, versioned execution artifacts---and Req\,B as \emph{conversational flexibility} (\cf)---how directly natural-language or agentic interaction determines actions.
Whether existing systems also deliver the boundary properties depends on their architectural commitments and is examined in \S\ref{sec:discussion}.

% ---------- 3. Execution Authority and Architectural Paradigms ----------
\section{Execution Authority and Architectural Paradigms for Conversational Scientific Workflows}
\label{sec:exec-authority}
The \ed/\cf axes provide a framework for comparing systems, but applying them requires a lens that distinguishes \emph{how} different systems mediate the transition from intent to computation. We capture this through \textit{execution authority}: the component that determines the concrete executable behavior contributing to the scientific record. Operationally, we identify it by the system's \textit{final execution artifact}---the last machine-consumable specification that the runtime accepts as ``what to run'' (e.g., a generated script, a schema-validated invocation, or a workflow specification). This is distinct from \textit{conversational authority}, which governs how the system interprets intent, proposes actions, and explains choices.

Using the \ed and \cf axes (\S\ref{sec:req-to-axes}), this section reviews 20 representative systems through the lens of execution authority, scoring each on the ordinal rubric in Table~\ref{tab:table2}, characterising five groups along a validation-scope spectrum, and visualising them in an \ed/\cf design space.

% ---------- 3.1 ----------
\subsection{Review scope and classification scheme}
\label{sec:survey-scope}
Table~\ref{tab:table1} summarizes 20 representative systems---commercial products, open-source frameworks, and canonical workflow systems. Systems were selected for: (i)~coverage of all five validation-scope groups, (ii)~public documentation sufficient for evidence-based scoring, (iii)~an established user base or significant community adoption, and (iv)~architecturally distinct profiles within each group (e.g., Galaxy and Nextflow differ in execution model and community norms). The selection is illustrative rather than exhaustive; systems are classified along five attributes: \emph{executable unit}, \emph{final execution artifact}, \emph{validation before execution}, \emph{provenance capture}, and \emph{primary execution authority}. We include ``generic use'' of general-purpose LLM chat as a baseline. Each system is also assigned ordinal \ed and \cf scores (rubric in Table~\ref{tab:table2}); the resulting design-space plot (Figure~\ref{fig:figure2}) is discussed in \S\ref{sec:design-space}.

Based on these attributes, each system is assigned to one of five groups along a \emph{validation-scope spectrum} (reflecting typical/default usage; specific deployments may differ): \textbf{Generative} (no formal pre-execution gate), \textbf{Tool-augmented} (tool-level validation only), \textbf{Schema-gated} (chat-first with mandatory schema-level validation), \textbf{Workflow + NL} (workflow-first with natural-language (NL) interfaces), and \textbf{Workflow-centric} (explicit workflow artifact with strong static validation). Row colours in Table~\ref{tab:table1} encode these groups; the table also uses the standard abbreviations VCS (version control system) and CI (continuous integration).

\paragraph{Ordinal scoring for the design-space plot (ED/CF).}
\label{sec:ordinal-scoring}
Each system is assigned ordinal (1--5) \ed and \cf scores using the rubric in Table~\ref{tab:table2}, visualised in Figure~\ref{fig:figure2}.

% ---------- Table 2: ED/CF Rubric ----------
\begin{table}[htbp]
  \centering
  \small
  \caption{Ordinal rubric for execution determinism (\ed) and conversational flexibility (\cf).}
  \label{tab:table2}
  \setlength{\tabcolsep}{6pt}
  \renewcommand{\arraystretch}{1.6}
  \begin{tabularx}{\linewidth}{@{}c X X@{}}
    \toprule
    \textbf{Score} & \textbf{\ed{} (Execution Determinism)} & \textbf{\cf{} (Conversational Flexibility)} \\
    \midrule
    \rowcolor{gray!6}
    1 & Unconstrained generated code; no formal pre-execution gate & No conversational interface (DSL/config/GUI only) \\
    2 & Primarily runtime checks; weak pre-execution constraints & Limited NL assistance (form help, search, hints); NL does not control execution \\
    \rowcolor{gray!6}
    3 & Developer-defined tool interfaces; partial validation but not uniformly schema-governed & NL assists authoring/modifying workflows; execution requires an externalized artifact \\
    4 & Schema/type-validated tool calls with strong pre-execution validation & NL directly selects/parameterizes tool/workflow invocations via structured calls \\
    \rowcolor{gray!6}
    5 & Explicit workflow spec/registry with static validation; reproducibility by construction & Free-form NL or agentic loops directly driving actions with minimal structural constraint \\
    \bottomrule
  \end{tabularx}
  \normalsize
\end{table}

% ---------- Table 1: Full Survey Table (shortened text) ----------
\begin{table}[htbp]
\centering
\scriptsize
\setlength{\tabcolsep}{4pt}
\renewcommand{\arraystretch}{1.2}
\hyphenpenalty=10000
\exhyphenpenalty=10000
\caption{Reviewed systems and architectural attributes (IDs match Fig.~\ref{fig:figure2}). Row colour encodes paradigm group: red~= generative, orange~= tool-augmented, green~= schema-gated, purple~= workflow + NL, blue~= workflow-centric. \ed/\cf{} scores are reported in Figure~\ref{fig:figure2} and Appendix~\ref{app:scoring-protocol}.}
\label{tab:table1}
\adjustbox{max width=\textwidth}{%
\begin{tabular}{@{}p{0.5cm}>{\raggedright}p{2.6cm}p{1.6cm}p{2.2cm}p{2.4cm}p{2.2cm}p{2.2cm}@{}}
\toprule
\textbf{ID} & \textbf{System} & \textbf{Domain} & \textbf{Interface} & \textbf{Executable} & \textbf{Validation} & \textbf{Provenance}\\
\midrule
\rowcolor{grpGenerative!10}
1  & \textit{LLM chat (generic)} & General & Chat (free-form) & Generated code & None/informal & Ad hoc logs\\
\rowcolor{grpGenerative!10}
2  & \textit{GitHub Copilot} & Software dev & Inline prompts & Code snippets & None & External (VCS/CI)\\
\rowcolor{grpGenerative!10}
3  & \textit{AutoGPT-style agents} & General & Autonomous agent & Plans + code + tools & Runtime-only & Partial logs\\
\rowcolor{grpGenerative!10}
4  & \textit{ReAct-style frameworks} & General & Reason--act loops & Plans + tool actions & Tool checks & Partial logs\\
\addlinespace[3pt]
\rowcolor{grpToolAug!10}
5  & \textit{LangChain} & LLM framework & Chat + agents & Code-defined tools & Limited checks & Partial (obs/logs)\\
\rowcolor{grpToolAug!10}
6  & \textit{Multi-agent science} & Discovery & Conv.\ agents & Tool graphs/plans & Partial checks & Partial logs\\
\rowcolor{grpToolAug!10}
7  & \textit{Semantic Kernel} & LLM framework & Chat + plugins & Tool invocations & Partial (dev types) & Partial (telemetry)\\
\addlinespace[3pt]
\rowcolor{grpAgentic!10}
8  & \textit{OpenAI Assistants / tool-calling} & General & Chat + tool calls & Tool invocations & Strong if schema & Partial (app logs)\\
\rowcolor{grpAgentic!10}
9  & \textit{Copilot Studio / Power Automate} & Enterprise auto & Chat + low-code & Connectors + flows & Strong (contracts) & Explicit (run hist.)\\
\addlinespace[3pt]
\rowcolor{grpWfNL!10}
10 & \textit{n8n} & Automation & Low-code (+NL) & Node workflows & Strong (node schema) & Explicit (runs)\\
\rowcolor{grpWfNL!10}
11 & \textit{NL-assisted wf authoring} & Sci workflows & NL-assisted & Gen.\ $\rightarrow$ validated wf & Strong (post-val.) & Explicit (after exp.)\\
\rowcolor{grpWfNL!10}
12 & \textit{Dataiku DSS} & Analytics/ML & GUI (+NL) & Recipes/workflows & Strong (typed) & Explicit (lineage)\\
\addlinespace[3pt]
\rowcolor{grpWfCentric!10}
13 & \textit{Galaxy} & Bioinfo & Forms (ltd.\ NL) & Workflows (DAG) & Strong & Explicit (histories)\\
\rowcolor{grpWfCentric!10}
14 & \textit{Snakemake} & Sci workflows & DSL/config & Workflows (DAG) & Strong & Explicit (files/logs)\\
\rowcolor{grpWfCentric!10}
15 & \textit{Nextflow} & Sci workflows & DSL/config & Workflows (DAG) & Strong & Explicit (runs/meta)\\
\rowcolor{grpWfCentric!10}
16 & \textit{nf-core} & Bioinfo & Templates & Curated workflows & Strong (CI+valid.) & Explicit (versioned)\\
\rowcolor{grpWfCentric!10}
17 & \textit{Workflow registries} & Multi-domain & Registry/portal & Registered workflows & Strong (policy) & Explicit (versioned)\\
\rowcolor{grpWfCentric!10}
18 & \textit{AWS Step Functions} & Cloud wf & DSL + console & State-machine wf & Strong (static) & Explicit (history)\\
\rowcolor{grpWfCentric!10}
19 & \textit{Apache Airflow} & Data/ML & Python DAG + UI & DAGs & Strong (DAG parse) & Explicit (DB+logs)\\
\rowcolor{grpWfCentric!10}
20 & \textit{Kubeflow / Argo} & ML wf & UI + YAML/SDK & Pipelines (DAG) & Strong (spec) & Explicit (artifacts)\\
\bottomrule
\end{tabular}%
}
\end{table}

\paragraph{Scoring protocol.}
Each of the 20~systems was assessed in 15~independent scoring sessions across three model families---ChatGPT~5.2, Claude Sonnet~4.6, and Gemini~3.1~Pro---each in a fresh conversational context using the same structured prompt. Across all 15~runs Krippendorff's $\alpha$ was 0.80 for \ed{} and 0.98 for \cf{}, indicating substantial-to-near-perfect inter-model agreement. Figure~\ref{fig:figure2} uses 15-run means to separate systems that share the same median position. Per-family consistency statistics, the full prompt template, per-session scores, and inter-model agreement analysis are provided in Appendix~\ref{app:scoring-protocol}.

\noindent \ed captures the extent to which ``what runs'' is constrained to validated, replayable artifacts independent of conversational variability. It does not guarantee scientific correctness or eliminate nondeterminism inside individual tools (e.g., stochastic algorithms). The reviewed systems trace a Pareto front in the \ed/\cf{} space (Figure~\ref{fig:figure2}): no system simultaneously exceeds the frontier on both axes. This front is empirical but partly structural, since systems granting NL direct execution authority tend by construction to forgo pre-execution validation (\S\ref{sec:future-work}). Scores are ordinal positions for comparative visualisation, not quantitative performance measurements.

% ---------- 3.2 ----------
\subsection{Architectural design space}
\label{sec:design-space}
\begin{figure}[t]
  \centering
  \includegraphics[width=1\textwidth]{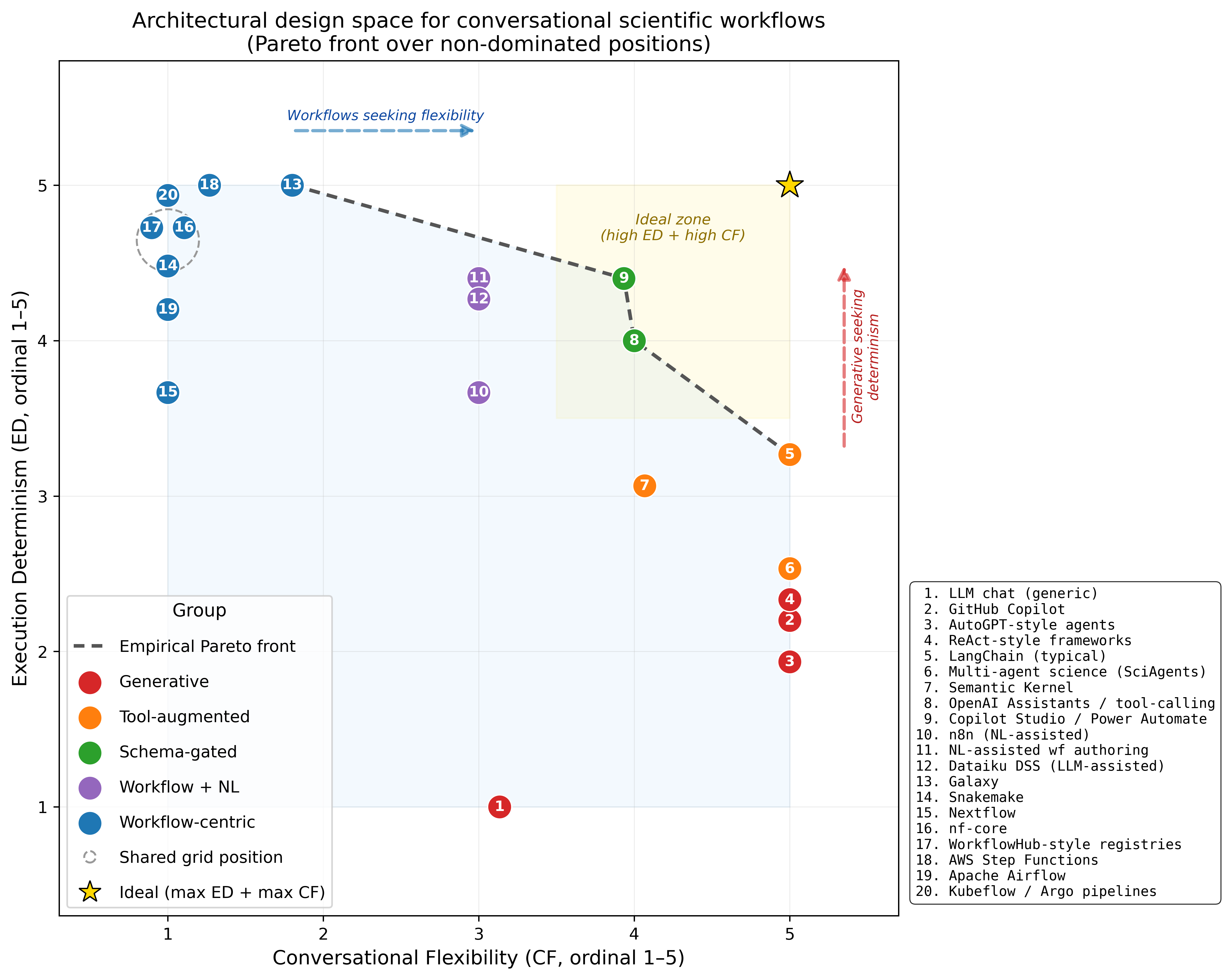}
  \caption{Architectural design space for conversational scientific workflows. Systems are positioned by mean \ed and \cf scores across 15~independent scoring runs (Appendix~\ref{app:scoring-protocol} reports consensus medians; means are used here for visual separation of co-located systems); point labels correspond to system IDs. Colours encode five groups: generative (red), tool-augmented (orange), schema-gated (green), workflow + NL (purple), and workflow-centric (blue). The dashed line is the empirical Pareto front connecting non-dominated positions with piecewise-linear segments (i.e.\ straight lines between observed Pareto-optimal points); the shaded region below the front contains dominated positions. The gold star marks the ideal region of high flexibility and high determinism. \emph{Caveat:} scores are ordinal (1--5) and non-quantitative; the piecewise-linear front is displayed for visual communication of the trade-off shape and should not be interpreted as implying interval-scale distances between positions or a continuous functional relationship.}
  \label{fig:figure2}
\end{figure}

Figure~\ref{fig:figure2} visualises the reviewed systems in the \ed/\cf space. These axes are properties of system architecture, not of language models; the same model can sit anywhere depending on how execution is mediated. The spatial separation of the five groups traces a non-linear trade-off captured by the empirical Pareto front.

The dashed line in Figure~\ref{fig:figure2} traces this front---piecewise-linear segments connecting non-dominated positions---showing the highest \ed observed for a given \cf level (or vice versa). Straight-line interpolation is used because the underlying scores are ordinal, to avoid implying a continuous functional relationship.

The \ed/\cf axes describe \emph{what practitioners require}---perfectly flexible interaction \emph{and} perfectly deterministic execution---not intrinsic software properties. The (5,\,5) ideal therefore represents a demand-side attractor: all systems face pressure to move toward it. Over time, generative tools will add validation and provenance while workflow-centric tools add conversational interfaces, so Figure~\ref{fig:figure2} should be read as a dated snapshot of an actively shifting landscape.

We define the \emph{schema-gated zone} as the region where both \ed{} and \cf{} reach or exceed~3.5. Of the 20~reviewed systems, only IDs~8 and~9 occupy this zone---both agentic systems that enforce mandatory schema validation on every tool invocation. The workflow + NL group (IDs~10--12) achieves strong \ed{}~($\geq 3.7$) but falls short on \cf{}~($= 3.0$), while the tool-augmented group (IDs~5--7) lacks the mandatory validation to elevate \ed{} above the threshold. Both groups are converging toward the zone but have not yet reached it.

Although the taxonomy identifies five groups, architecturally they cluster into three paradigms. The generative and tool-augmented groups share a trait---the LLM retains direct execution authority---and are examined together as \textbf{Paradigm~I} (\S\ref{sec:paradigm-generative}). The workflow-centric and workflow + NL groups both vest execution authority in an explicit workflow specification: \textbf{Paradigm~II} (\S\ref{sec:paradigm-workflow}). The schema-gated group (IDs~8--9) forms \textbf{Paradigm~III} (\S\ref{sec:pattern-schema-gated}): these systems already instantiate the pattern at the tool-call level; we propose that extending gating to the composed-workflow level can push toward simultaneously high \ed{} and \cf{}.

% ---------- 3.3 ----------
\subsection{Paradigm I: Generative code-centric approaches}
\label{sec:paradigm-generative}
The first region grants conversational reasoning direct authority over executable behavior: LLMs translate prompts into code, scripts, notebook cells, or shell commands (Table~\ref{tab:table1}, IDs 1--4; Figure~\ref{fig:figure2}). A closely related set of \textbf{tool-augmented} systems (IDs 5--7) add developer-defined tool interfaces---function-calling APIs, plugin registries, or protocol-based tool servers (e.g., the Model Context Protocol, MCP~\cite{Anthropic2024MCP})---that validate individual invocations against a schema, though validation remains limited to single calls. These systems trade formalisation for expressivity, suiting exploratory analysis and rapid prototyping, but coupling conversational and execution authority means generated code is often ad hoc, behaviour may vary across runs, and validation is informal. Widely adopted agentic code assistants---GitHub Copilot, Cursor, Claude Code---incorporate sophisticated internal planning and execution tracing, but these remain largely opaque to the user; for scientific R\&D, transparency, auditability, and human-in-the-loop control must be first-class, user-facing features.

\paragraph{Architectural signature.}
Execution authority resides in the LLM/agent (or its controller), and the final execution artifact is typically generated code or loosely constrained tool actions. The contract between intent and execution is primarily conversational rather than machine-checkable.

% ---------- 3.4 ----------
\subsection{Paradigm II: Workflow-centric systems}
\label{sec:paradigm-workflow}
A second region comprises systems that prioritise execution determinism by externalising computation into explicit workflow specifications---domain-specific languages (DSLs), configuration files, or graphical editors defining a workflow directed acyclic graph (DAG), data dependencies, and parameterisation (Table~\ref{tab:table1}, IDs 13--20; Figure~\ref{fig:figure2}). They provide strong reproducibility, provenance, and scalability guarantees, suiting stable pipelines and collaborative reuse. The principal limitation is interaction cost: users must learn workflow abstractions and configuration syntax, and iterative experimentation often requires manual specification edits.

A closely related set of \textbf{workflow + NL} platforms (IDs~10--12) augment this pattern with NL interfaces---conversational authoring, NL-to-workflow translators, or chat-based parameter completion---while retaining the validated workflow specification as execution authority. They achieve strong \ed{}~($\geq 3.7$) but limited \cf{}~($= 3.0$), placing them outside the schema-gated zone (\S\ref{sec:design-space}).

\paragraph{Architectural signature.}
Execution authority resides in the workflow specification (or a curated workflow registry/executor), and the final execution artifact is an explicit workflow instance (DAG/DSL/config) that the runtime validates and executes. Natural language may assist the human author but does not control execution.

\subsection{Paradigm III: Schema-gated tool execution}
\label{sec:pattern-schema-gated}
The third region contains the \emph{schema-gated} group (Table~\ref{tab:table1}, IDs 8--9; Figure~\ref{fig:figure2}): chat-first, agentic systems that enforce mandatory schema validation on every tool invocation. Unlike generative systems (which let the LLM execute freely) or workflow-centric systems (which require users to author specifications), these retain full conversational interaction while mediating execution through validated, structured artifacts.

Of the 20~reviewed systems, IDs~8 and~9 sit closest to the (5,\,5) ideal---the only systems inside the schema-gated zone defined in \S\ref{sec:design-space}. What places them there is mandatory, gateway-style schema validation on every tool invocation:

\begin{itemize}
  \item \textbf{OpenAI Assistants / tool-calling (ID~8).} In \texttt{strict:\,true} mode, the Assistants API provides strong schema-level validation via constrained decoding, rejecting function-call argument sets that do not conform to a developer-defined JSON Schema before dispatch. The LLM reasons freely about which tool to call and how to fill its parameters (CF$\,=4$), but no tool executes unless the schema gate passes (ED$\,=4$).
  \item \textbf{Copilot Studio / Power Automate (ID~9).} Connectors are defined via OpenAPI specifications with typed parameters and operation contracts; Power Automate's Flow Checker validates a flow against connector schemas before execution, and each run produces an explicit history entry. A generative orchestration layer selects and sequences topics, tools, and agents from user messages (CF$\,=4.4$), but every connector invocation must satisfy its contract (ED$\,=3.9$).
\end{itemize}

\noindent In both cases, validation is \emph{mandatory and exclusive}: there is no execution path that bypasses the schema boundary. This is the property that distinguishes these systems from tool-augmented systems (IDs~5--7), which offer developer-defined tool interfaces but permit the LLM to reason its way into unvalidated actions. Schema validation of individual tool calls is already widespread---OpenAI function calling, Anthropic tool use, and MCP all define typed interfaces for single invocations---but IDs~8 and~9 go further by making the schema the sole path to execution.

\paragraph{Architectural signature.}
Execution authority resides in a schema-validated tool/workflow layer; the final execution artifact is a fully resolved, machine-checkable invocation object. Table~\ref{tab:table3} summarises the paradigm distinctions. Sections~\ref{sec:architecture}--\ref{sec:ref-arch} formalise the underlying principle and extend it from tool-level gating to composed-workflow orchestration.

% ---------- Table 3 ----------
% NOTE: This is the *small summary table* (keep as authored text; not CSV-backed).
\begin{table}[htbp]
  \centering
  \small
  \caption{Architectural paradigms for conversational scientific workflows (summary). Column colours correspond to the paradigm groups defined in Table~\ref{tab:table1}.}
  \label{tab:table3}
  \setlength{\tabcolsep}{6pt}
  \renewcommand{\arraystretch}{1.6}
  \raggedright
  \begin{tabularx}{\linewidth}{@{}p{3.6cm}>{\columncolor{grpGenerative!10}}X>{\columncolor{grpWfCentric!10}}X>{\columncolor{grpAgentic!10}}X@{}}
    \toprule
    \textbf{Dimension} & \cellcolor{grpGenerative!25}\textbf{Generative} & \cellcolor{grpWfCentric!25}\textbf{Workflow-centric} & \cellcolor{grpAgentic!25}\textbf{Schema-gated}\\
    \midrule
    \addlinespace[2pt]
    \emph{Primary execution authority} & LLM/Agent & Workflow spec / registry & Schema-validated tool/workflow layer\\
    \addlinespace[2pt]
    \emph{Role of natural language} & Produces executable artifacts & Assists authoring (limited) & Expresses intent; resolves parameters; proposes actions\\
    \addlinespace[2pt]
    \emph{Executable unit} & Generated scripts/cells/commands & Explicit DAG/DSL/config & Validated tool calls and workflows\\
    \addlinespace[2pt]
    \emph{Validation before execution} & Minimal or absent & Strong & Strong (schema/type/constraint checks)\\
    \addlinespace[2pt]
    \emph{Determinism across runs} & Not guaranteed & High & High (within validated bounds)\\
    \addlinespace[2pt]
    \emph{Provenance capture} & Implicit / ad hoc & Explicit & Explicit (tool- and workflow-level)\\
    \addlinespace[2pt]
    \emph{Iterative exploration} & High, but unsafe & Limited / high overhead & High, within governed constraints\\
    \addlinespace[2pt]
    \emph{Typical failure modes} & Hallucinated APIs, silent variation & Configuration friction, rigidity & Coverage limits of schema/tool registry\\
    \bottomrule
  \end{tabularx}
  \normalsize
\end{table}

% ---------- 4. Schema-Gated Orchestration ----------
\section{Schema-Gated Orchestration as a Design Principle}
\label{sec:architecture}
The two systems in the schema-gated group (IDs~8--9, \S\ref{sec:pattern-schema-gated}) share an architectural commitment that warrants formalisation. We use \textit{schema-gated} to mean that \textbf{the runtime refuses to execute unless the proposed action can be represented as a versioned tool or workflow invocation that validates against a machine-checkable schema at the boundary}. Conversation may shape intent, but cannot bypass validation. This makes a separation-of-authority decision explicit:
\begin{itemize}
  \item \textbf{Conversational authority:} interpreting intent, proposing candidate actions, asking clarifying questions, and explaining choices.
  \item \textbf{Execution authority:} selecting and running computation only through actions that satisfy explicit, machine-readable constraints.
\end{itemize}
Operationally, this makes ``what ran'' an explicit object---a tool/workflow identifier (and version) plus a resolved parameter set---rather than an implicit prompt outcome. A system that uses schemas to \emph{describe} tool interfaces but permits unvalidated actions is schema-\emph{described}; a schema-gated system makes the schema the sole path to execution.

IDs~8 and~9 already enforce this invariant at the individual tool-call level. However, scientific workflows typically chain multiple steps with cross-step dependencies that single-call validation cannot catch. Consider a materials-discovery pipeline: (i)~load a dataset, (ii)~train a surrogate model on columns from that dataset, and (iii)~run inverse design using the trained model. Tool-level validation can confirm each call is individually well-typed, yet cannot detect that the target columns in step~(ii) do not exist in step~(i)'s dataset, or that step~(ii)'s model was trained on different properties than step~(iii) requires. These are \emph{cross-step} type and dependency errors: each invocation is locally valid, but the composed plan is unsound. A workflow-level schema gate catches such mismatches before any computation runs by validating inter-step data-flow types, dependency ordering, and parameter compatibility across the entire DAG.

For the determinism that scientific practice demands (\ed{}$\,=5$), the schema gate must extend from individual tool calls to the composed-workflow level, enforcing structural, dependency, and type constraints across the entire plan before any step executes. We call the tool-level pattern \emph{schema-gated tool execution} and the workflow-level extension \emph{schema-gated orchestration}.

\subsection{Operational principles}
\label{sec:design-principles}
The schema-gated invariant---nothing executes without schema validation---carries operational consequences that distinguish it from both the unconstrained agentic systems and the rigid workflow systems reviewed above.

\paragraph{Clarification-before-execution.}
Because the schema gate rejects incomplete or ill-typed invocations, missing fields, type mismatches, and constraint violations surface as conversational prompts---turning silent failures into structured negotiation. The invocation object is also the artefact that approval gates examine and provenance records capture, making boundary properties (\S\ref{sec:requirements}) architectural consequences rather than bolted-on features.

\paragraph{Constrained plan--act orchestration.}
Schema-gated orchestration separates reasoning from execution. In \textbf{planning mode}, the model operates as a fully agentic reasoner---interpreting intent, decomposing tasks, and conducting parameter completion through open-ended dialogue~\cite{Yao2022}. In \textbf{action mode}, the schema-gated invariant applies: nothing executes without validation. An unconstrained agent \emph{can} skip validation and act on approximate reasoning; a schema-gated system \emph{cannot}. This two-mode structure enables high conversational flexibility in planning to coexist with high execution determinism in action.

\paragraph{From tool-level to workflow-level gating.}
Tool-level schema validation---OpenAI function calling, Anthropic tool-use, MCP---is already mainstream but applies only to individual calls. Schema-gated orchestration extends validation to composed workflows, enforcing structural, dependency, and type constraints across multi-step plans before execution. The key distinction is \emph{opaque} vs.\ \emph{transparent} composition: a pre-curated MCP tool wrapping several internal steps provides a defined interface but hides the internal sequence, whereas a schema-gated workflow exposes it as a validated DAG whose structure and inter-step types are machine-checkable (\S\ref{sec:validation-framework}). Users or LLMs can substitute a step, add a stage, or change a parameter, and the system re-validates the entire plan.

% ---------- 5. Reference Architecture ----------
\section{Reference Architecture}
\label{sec:ref-arch}
This section presents a reference architecture that instantiates schema-gated orchestration and its three operational principles (\S\ref{sec:architecture}), proposing a design that may decouple the \ed/\cf{} trade-off.

\subsection{Architecture overview}
\label{sec:arch-overview}
The reference architecture separates concerns into distinct layers with well-defined interfaces so that components can evolve independently (e.g., switching LLM providers without altering the execution engine). The key commitment is the separation between a schema-validated registry and a conversational layer, bridged by an orchestration controller that enforces validation and produces structured provenance. A chat-first UI exposes the schema-gated boundary directly, making validation states, execution progress, and clarification loops visible.

\subsection{Orchestration controller}
\label{sec:controller}
The orchestration controller mediates between NL interaction and validated execution. As conversation progresses, the controller accumulates context---chat history, prior tool calls, and returned results---from which the LLM selects the appropriate platform action at each turn. Execution is dispatched only once the user is satisfied with the proposed action \emph{and} all schema constraints are met.

Two kinds of callable operate at different levels. \emph{Platform actions}---\texttt{search\_workflows}, \texttt{get\_parameters}, \texttt{list\_datasets}, \texttt{execute\_workflow}---let the LLM navigate and operate the system. \emph{Domain tools} are the atomic scientific operations that compose into workflows. Both are schema-validated; platform actions govern how users discover and trigger work, domain tools govern what actually runs.

\begin{figure}[t]
  \centering
  \includegraphics[width=1\textwidth]{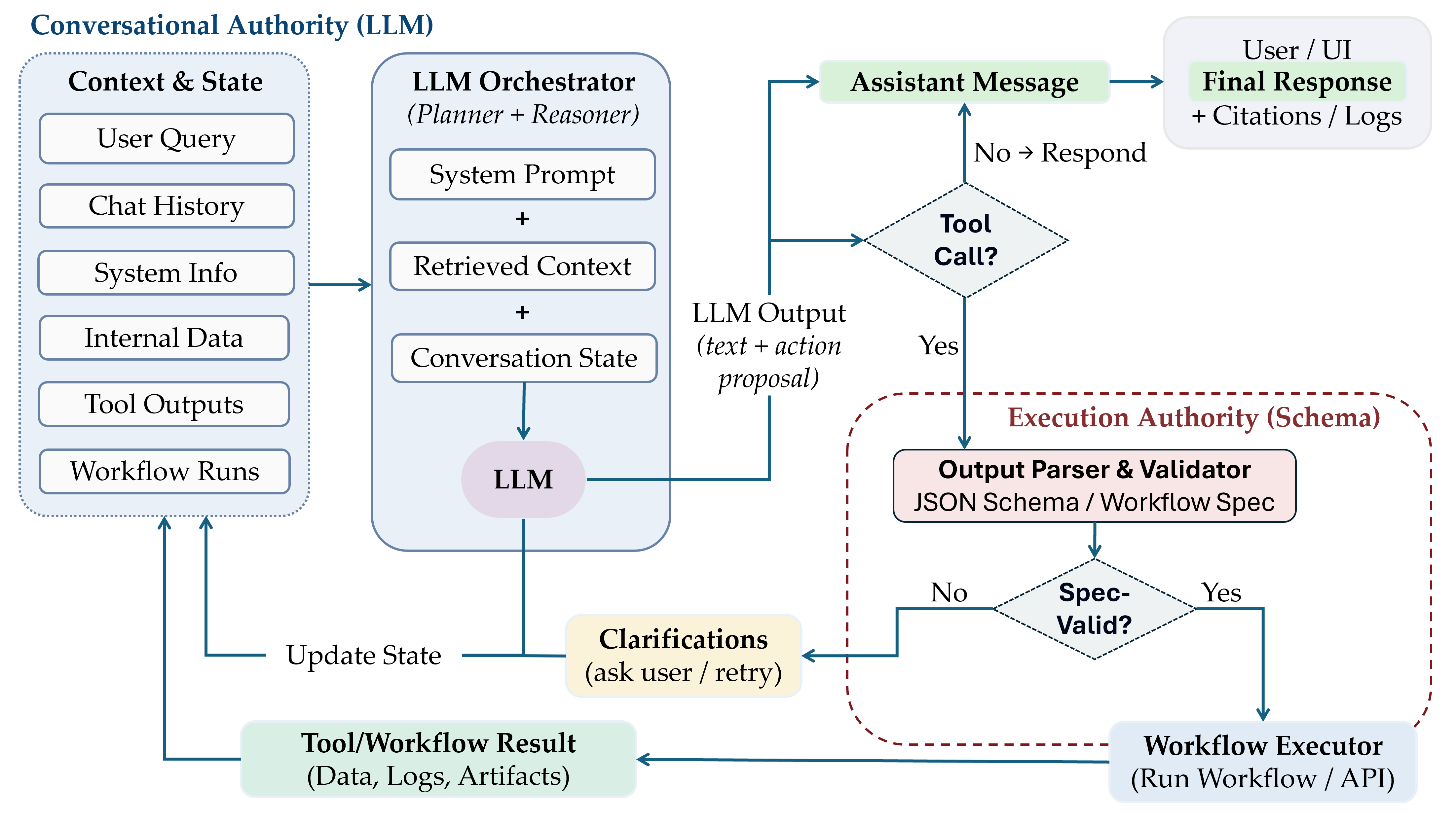}
  \caption{Reference architecture separating conversational authority from execution authority. Context and state (user query, chat history, system information, internal data, and prior tool/workflow outputs) are assembled for an LLM orchestrator (planner/reasoner), which produces an assistant message and, optionally, an action proposal. Proposals enter the execution-authority gate (red dashed region): an output parser validates them against a JSON schema or workflow specification. Invalid proposals trigger a clarification loop; valid proposals are forwarded to the workflow executor. Resulting data, logs, and artifacts update the shared context for subsequent turns. The LLM may converse freely, but cannot execute---execution authority resides entirely in schema validation.}
  \label{fig:figure3}
\end{figure}

The controller's scope is \emph{intent understanding and execution dispatch}, not workflow construction. The orchestrator consumes validated workflows from the registry; it does not create them during a session. The following invocation illustrates a platform action that dispatches a validated workflow:

\begin{verbatim}
{"tool": "execute_workflow",
 "workflow_id": "alloy_inverse_design",
 "parameters": {
   "dataset_id": "123e4567-e89b-12d3-a456-426614174000",
   "target_properties": ["yield_strength", "creep_life"],
   "constraints": {"Cr": {"max": 12.0}, "Co": {"min": 5.0}},
   "n_candidates": 50
 }}
\end{verbatim}
These invocation objects constrain the model's influence to auditable, schema-compliant actions. Figure~\ref{fig:figure3} illustrates the resulting architecture: the LLM orchestrator holds conversational authority while execution authority resides in schema validation, so that each action proposal must pass the execution-authority gate before any computation runs and each execution result feeds back into context for subsequent turns.

\subsection{Validation framework}
\label{sec:validation-framework}
Domain tools and workflows are declared as Python models that compile to machine-checkable schemas (e.g., JSON Schema) and are versioned alongside their implementations.

Because every domain tool declares typed inputs, outputs, and dependency constraints (\S\ref{sec:design-principles}), the dependency graph constrains the planning search space: given a target tool whose schema declares upstream requirements, the graph determines which components are needed and in what order---reducing composition from a combinatorial problem to a structurally guided traversal. Composition is therefore not limited to manual authoring: an LLM can propose candidate pipelines, a constraint solver can enumerate valid orderings, or a domain expert can assemble steps by hand---in each case the same schema-level checks validate the result. Existing workflows serve as reusable templates that users adapt by substitution or reparameterisation.

\subsubsection{Domain tool schema structure}
Domain tools define atomic operations such as data preprocessing, model training, or statistical analysis. A domain tool schema resembles an MCP tool definition---typed inputs, outputs, and a callable interface---but carries richer metadata: dependency declarations, provenance information (origin, version, maintainer), domain tags, and typed inter-step contracts. This metadata enables composition into workflows: because each tool advertises its input/output types alongside domain context, an orchestrator can match compatible tools rather than relying on manual wiring. A domain tool cannot enter the registry until it passes automated checks for parameter consistency, documentation completeness, and service availability (Figure~\ref{fig:figure4}). A full example is provided in Appendix~\ref{app:tool-schema}.

\begin{figure}[t]
  \centering
  \includegraphics[width=0.85\textwidth]{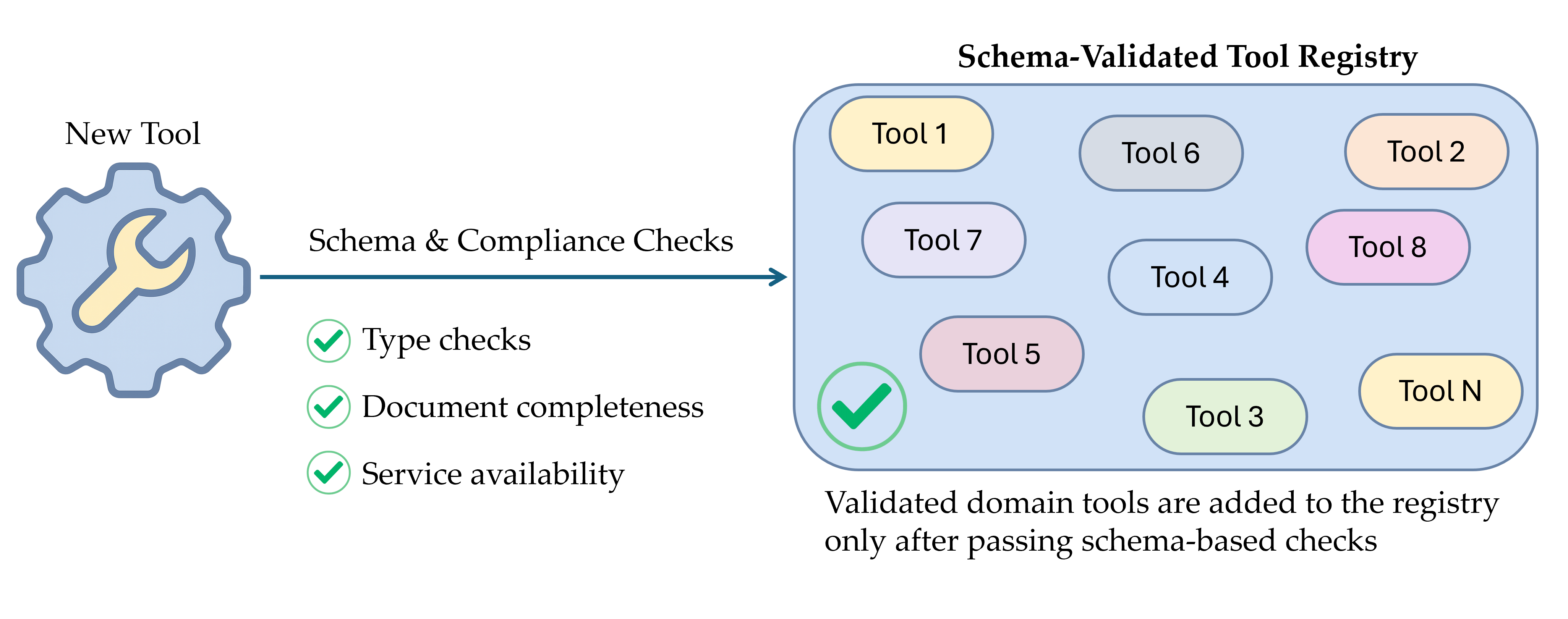}
  \caption{Domain tool schema validation and inclusion in the registry. New domain tools undergo schema and compliance checks---including type validation, documentation completeness, and service availability---before being added to the validated tool registry. Only schema-verified domain tools become available for workflow execution.}
  \label{fig:figure4}
\end{figure}

\subsubsection{Workflow schema structure}
Workflows compose domain tools into DAGs specifying execution order, dependencies, and data flow. A workflow schema defines the steps, the tools they invoke, inter-step dependencies, and parameter mappings. Schemas also capture workflow-level metadata (e.g., tags, complexity ratings) for discovery. Before execution, the engine validates graph acyclicity, inter-step type compatibility, parameter resolution, and tool availability (Figure~\ref{fig:figure5}). An extended example is in Appendix~\ref{app:workflow-schema}.

\begin{figure}[htbp]
  \centering
  \includegraphics[width=0.85\textwidth]{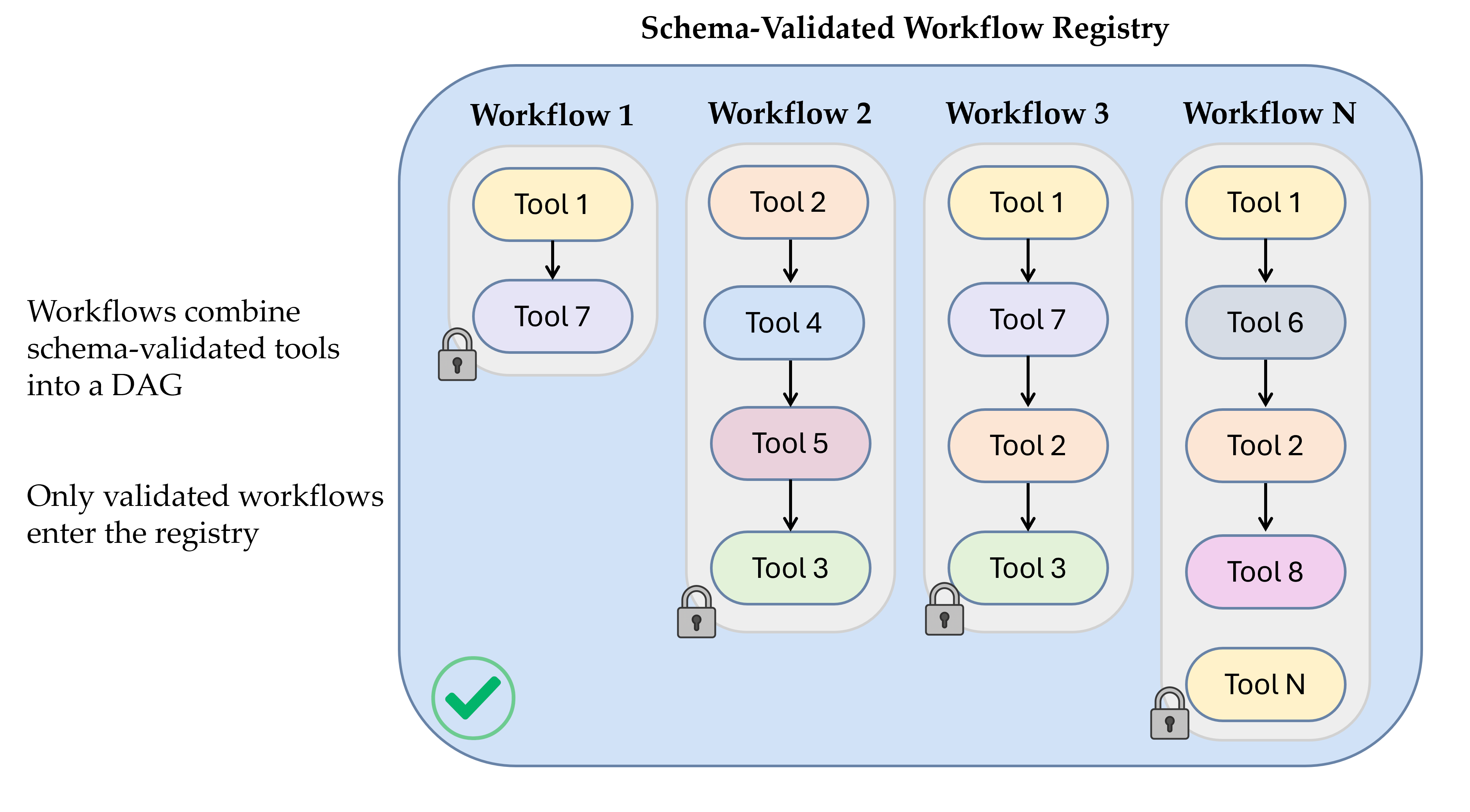}
  \caption{Workflow schema composition in the validated registry. Each workflow combines schema-validated domain tools into a DAG, defining explicit data and parameter flow between steps. Only workflows that pass acyclicity, type-compatibility, and parameter-resolution checks are included in the validated workflow registry.}
  \label{fig:figure5}
\end{figure}

\subsubsection{Schema-gated execution}
\begin{figure}[htbp]
  \centering
  \includegraphics[width=0.7\textwidth]{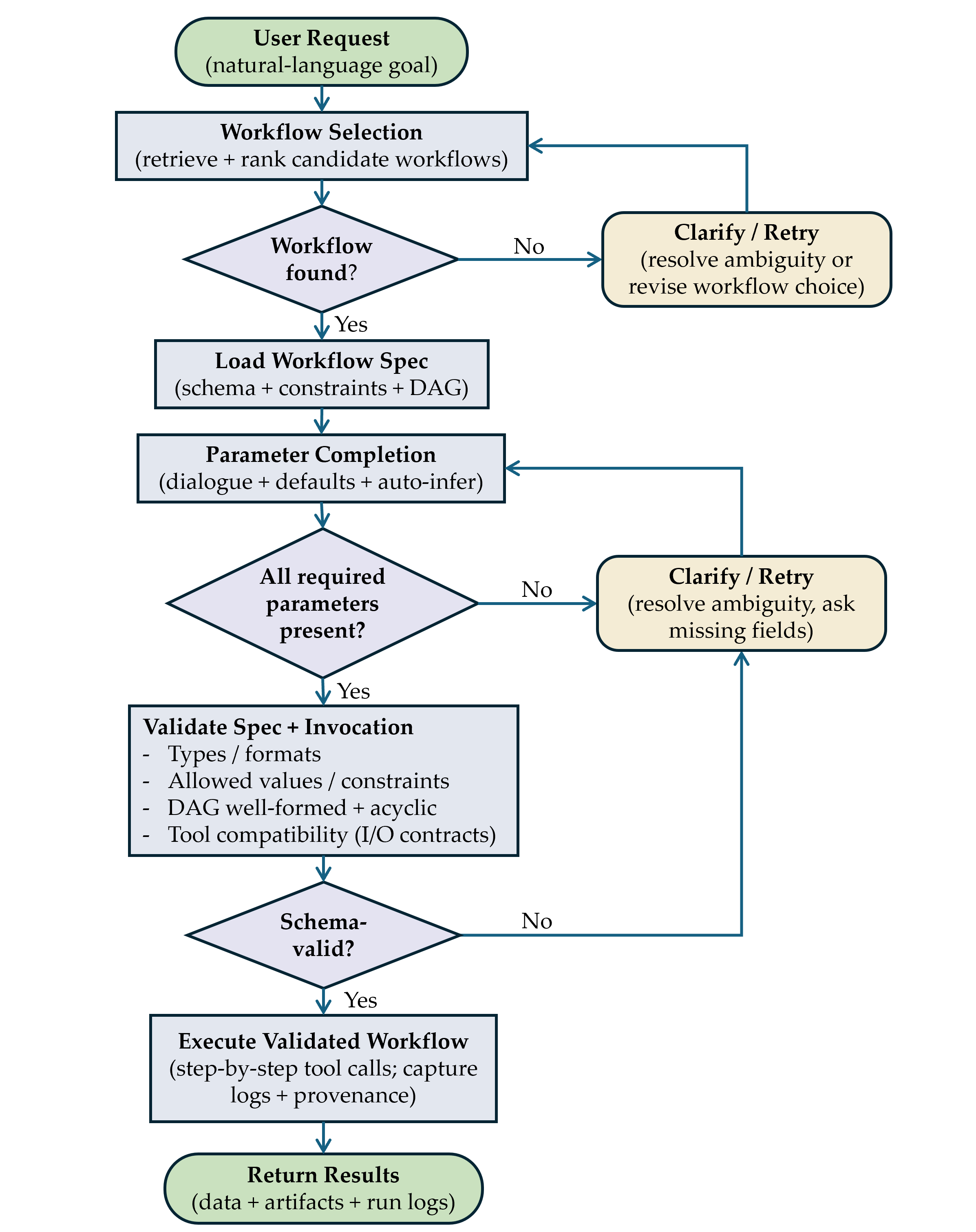}
  \caption{Schema-gated workflow execution process. The system identifies candidate workflows, loads the corresponding schema, collects and validates parameters through dialogue, and executes only after all constraints are satisfied. Validation failures trigger clarification loops; conversational context is updated throughout.}
  \label{fig:figure6}
\end{figure}

When a user selects (or is recommended) a workflow, the controller loads the corresponding schema to determine which parameters must be gathered through conversation. Execution does not begin until required inputs are supplied and validated for type correctness, admissible values, and structural consistency (Figure~\ref{fig:figure6}).

\subsection{Execution engine and provenance}
\label{sec:engine-provenance}
The execution engine runs validated workflows asynchronously and creates a provenance record for each run---capturing the workflow definition snapshot, resolved parameters, timestamps, environment metadata, and any failures. Because inputs and outputs are schema-typed, records are structured and queryable, supporting comparison across runs and long-term reproducibility.

\subsection{Schema-gated orchestration in practice}
\label{sec:practice}
The following interaction trace illustrates the schema-gated pattern end-to-end.

\begin{table}[!htbp]
  \centering
  \scriptsize
  \caption{Schema-gated orchestration in practice (interaction trace). Each row shows one turn in a representative dialogue; the \emph{Schema Gate} column identifies the validation checkpoint enforced before the system proceeds. Row colour distinguishes actors: blue~= user, green~= system.}
  \label{tab:interaction-trace}
  \setlength{\tabcolsep}{4pt}
  \renewcommand{\arraystretch}{1.6}
  \adjustbox{max width=\textwidth}{%
  \begin{tabularx}{\textwidth}{@{}c >{\raggedright}p{0.9cm} >{\raggedright}p{2.8cm} X >{\raggedright\arraybackslash}p{2.5cm}@{}}
    \toprule
    \textbf{Turn} & \textbf{Actor} & \textbf{Action / Tool Call} & \textbf{Detail} & \textbf{Schema Gate} \\
    \midrule
    \rowcolor{grpWfCentric!8}
    1 & User & NL request & ``I want to design a new superalloy with low chromium\ldots'' & --- \\
    \rowcolor{grpAgentic!8}
    2 & System & \texttt{search\_workflows} & Registry lookup returns ranked candidate workflows & Tool schema validates query params \\
    \rowcolor{grpWfCentric!8}
    3 & User & Selects workflow & Picks ``Alloy Design Pipeline v2.1'' from results & --- \\
    \rowcolor{grpAgentic!8}
    4 & System & \texttt{get\_parameters} & Retrieves workflow parameter schema (required fields, types, ranges) & Tool schema validates workflow ID \\
    \rowcolor{grpAgentic!8}
    5 & System & Parameter prompt & Surfaces required fields: \texttt{model\_id}, \texttt{target\_properties}, \texttt{constraints} & --- \\
    \rowcolor{grpWfCentric!8}
    6 & User & Supplies params & Provides model ID, targets, and composition constraints & --- \\
    \rowcolor{grpAgentic!8}
    7 & System & Validation & Checks type correctness, admissible ranges, structural consistency & Workflow schema enforces all constraints \\
    \rowcolor{grpAgentic!8}
    8 & System & \texttt{execute\_workflow} & Dispatches validated invocation; streams progress and results & Blocked until gate~7 passes \\
    \midrule
    \multicolumn{5}{@{}l}{\emph{Results feed back into conversational context}} \\
    \midrule
    \rowcolor{grpAgentic!8}
    9 & System & Results summary & Returns metrics ($R^2$, RMSE) and top candidate compositions into dialogue & --- \\
    \rowcolor{grpWfCentric!8}
    10 & User & Iterative refinement & ``Try leave-one-out instead of 5-fold'' & --- \\
    \rowcolor{grpAgentic!8}
    11 & System & Parameter update & Re-uses prior invocation with one changed parameter (\texttt{validation\_strategy}) & Workflow schema re-validates modified invocation \\
    \rowcolor{grpAgentic!8}
    12 & System & \texttt{execute\_workflow} & Dispatches variant invocation; both runs are directly comparable & Blocked until gate~11 passes \\
    \bottomrule
  \end{tabularx}%
  }
\end{table}

Table~\ref{tab:interaction-trace} illustrates the core loop. A user provides a natural-language request; rather than generating code, the system invokes a platform action (\texttt{search\_workflows}) and returns the structured result into the dialogue. This is the first boundary: even ``figuring out what to run'' is mediated by a validated platform action rather than implicit reasoning.

Once a workflow is chosen, \texttt{get\_parameters} retrieves the parameter schema and conversation fills in required fields. Execution is blocked until the invocation is complete and type-valid; missing or ill-typed values trigger clarification (\S\ref{sec:design-principles}).

This boundary also supports rapid iteration: edits are changes to explicit parameters rather than changes to generated code, so each run is a concrete invocation that can be compared directly. Turns~9--12 illustrate this: when the user says ``try leave-one-out instead of 5-fold,'' the system re-uses the prior invocation with a single changed parameter, enabling systematic ablation without code inspection.

% ---------- 5. Discussion ----------
\section{Discussion}
\label{sec:discussion}
We organise the discussion around what schema-gated execution enables, its costs and limits, governance implications, and open questions.

\subsection{What schema-gated execution enables}
\paragraph{Why schema-gating is positioned to decouple the trade-off.}
The empirical Pareto front (Figure~\ref{fig:figure2}) arises because reviewed systems couple conversational and execution authority: granting NL direct control over what runs tends to forgo pre-execution validation, while enforcing validation tends to restrict conversation. Schema-gated orchestration decouples these axes. In planning mode, the model reasons without constraint; in action mode, nothing executes without schema validation. The architecture thus inherits the \ed strengths of workflow-centric systems (explicit, validated, versioned artifacts) while preserving the \cf strengths of chat-first systems (NL-driven discovery, parameter completion, iterative refinement). The residual constraint is that interaction is guided toward operations the registry can represent---precluding the unconstrained generation that the rubric reserves for \cf{}$\,=5$ (Table~\ref{tab:table2})---but this is precisely the constraint that makes execution determinism possible.

Whether a deployed schema-gated system achieves this decoupling in practice---and in particular whether conversational friction, coverage gaps, and registry maintenance costs erode either axis under real workloads---is an open empirical question. We flag deployment-level measurement (coverage-gap frequency, parameter-completion latency, task-completion rates) as the most pressing near-term priority (\S\ref{sec:future-work}).

\paragraph{Boundary properties as architectural by-products.}
Because the front reflects coupling of conversational and execution authority rather than a theoretical bound, the decoupling makes the boundary properties identified in \S\ref{sec:requirements}---human approval and provenance---architectural by-products of the invocation object rather than features that must be engineered separately.

\subsection{Costs and limits}
The central cost is coverage. A schema-gated system can only execute what its registry represents; absent tools or workflows force refusal or transition to an explicit authoring path. This shifts effort from prompt engineering to registry maintenance---schema design, validation, versioning, and backward-compatible evolution---an organisational commitment most acute at domain boundaries. Well-established workflows amortise overhead over many users; frontier research may lack coverage precisely where flexibility matters most.

Three concrete near-term mitigation strategies can soften this hard boundary without abandoning the schema-gated invariant: (i)~\emph{template-based bootstrapping}, in which validated workflows serve as composable templates that users adapt by tool substitution or reparameterisation rather than authoring from scratch; (ii)~\emph{LLM-assisted schema drafting}, in which the model generates a conformant \texttt{ToolDefinition} from existing code, documentation, or function signatures, subject to the same automated validation gates applied to manually authored definitions (Appendix~\ref{app:tool-schema}); and (iii)~\emph{tiered registries} with differentiated trust levels, permitting community-contributed tools to coexist with curated tools under explicit provenance and review policies. Whether hybrid modes that additionally permit sandboxed generative execution can preserve meaningful guarantees remains an open question (\S\ref{sec:future-work}).

Longer term, the most promising path is a federated ecosystem: if protocols such as MCP~\cite{Anthropic2024MCP} were extended with inter-tool dependency declarations and typed data contracts, providers could publish schema-conformant tool packages that an orchestrator composes into validated cross-provider workflows. Coverage would then scale with the community rather than with any single team's registry (\S\ref{sec:future-work}).

Three further limits apply. First, clarification-before-execution introduces conversational friction: structured negotiation before every execution could slow exploration in high-parameter workflows. This can be mitigated by pre-populating fields from metadata or prior runs and by carrying forward previously resolved parameters across iterations (Table~\ref{tab:interaction-trace}, turns~10--12). Second, schemas enforce well-typed parameters and structural validity but cannot ensure scientific appropriateness---user judgment about assumptions, data quality, and model validity remains essential. Third, schema-gating provides architectural determinism at the boundary; end-to-end determinism additionally requires controlled seeds, containers, and pinned dependencies.

\subsection{Governance and safety implications}
Routing all actions through auditable, schema-validated interfaces eliminates a major class of governance risk---arbitrary code execution via conversation---but residual risks remain: dangerous registered tools, prompt injection targeting tool calls, and data leakage.

The interview data underscore these concerns. Security \& IP Protection was the fourth-most-discussed theme (67~mentions, 9/10 stakeholders; Table~\ref{tab:thematic-groups}), with practitioners raising confidentiality of proprietary formulations, export-control constraints, and the risk that conversational context could expose trade secrets to cloud-hosted LLMs. Schema-gated orchestration addresses several concerns structurally: every action is mediated by a validated invocation object, enabling role-based access control at the tool and parameter level, data-classification tags that prevent sensitive values from reaching external services, and tamper-evident audit trails mapping each execution to a versioned schema and authenticated identity.

Nevertheless, schema-gating alone is insufficient. Defence in depth remains essential: least-privilege tool design, approval gates for sensitive actions, input sanitisation against prompt injection, and network-level isolation for on-premise deployments. The governance of the registry itself---who may publish, review, and retire tools---is a policy design problem mirroring open-source package ecosystem challenges.

\subsection{Design guidance and open questions}
The review and requirements suggest practical guidance for building trustworthy conversational workflow systems:
\begin{itemize}
  \item Treat the \textit{validated invocation} as the unit of execution, provenance, and approval.
  \item Make boundary failures explicit and conversationally actionable (missing fields, constraint violations, unsupported operations).
  \item Invest in registry lifecycle management (testing, versioning, schema evolution) as a first-class engineering practice.
  \item Use conversation to reduce interaction overhead (discovery and parameter completion), not to define executable behavior.
\end{itemize}

\noindent The thematic data reinforce these priorities: System Integration was the most discussed theme (191~mentions, 10/10~stakeholders), confirming connector coverage as a prerequisite; Workflow \& Task Automation and Data Search \& Retrieval follow at universal breadth. Narrow-reach themes such as Progressive Autonomy (19, 5/10) and Data Aggregation \& Anonymisation (10, 4/10) are emerging concerns to monitor but need not be prioritised at launch.

Several open questions remain: whether registries should be centrally curated, community-extensible, or tiered; how they should balance stability with rapid addition of new capabilities; and how systems should communicate uncertainty when registry coverage is lacking.

\subsection{Future work}
\label{sec:future-work}

An internal research testbed instantiating this architecture is under development. Several directions warrant investigation:
\begin{enumerate}[nosep,leftmargin=*]
  \item \emph{Broader stakeholder recruitment} beyond the Intellegens network, targeting organisations that rely on generative or ad hoc tooling.
  \item \emph{Federated tool ecosystems}: extending protocols such as MCP with inter-tool dependency declarations and typed data contracts, enabling software-as-a-service (SaaS) providers and research groups to publish vetted, schema-conformant tool packages that an orchestrator can automatically compose into cross-provider workflows---the primary direction for scaling registry coverage beyond any single organisation (\S\ref{sec:discussion}).
  \item \emph{Expanding the reviewed system set}, particularly mid-range systems (\cf{}$\,=2$--$4$) that could refine the Pareto front.
  \item \emph{Empirical evaluation of a deployed proof of concept}: schema authoring burden, coverage-gap frequency, conversational workflow friction, and user task-completion times---the most pressing near-term priority for measuring where a schema-gated system actually sits in the \ed/\cf{} space.
  \item \emph{Longitudinal re-assessment} of the \ed/\cf{} landscape at regular intervals, recording system versions, to track how the Pareto front evolves as platforms add new capabilities.
  \item \emph{Cross-domain interaction traces}: the interaction trace in Table~\ref{tab:interaction-trace} illustrates the pattern in a materials-science context; replicating it across additional domains would strengthen the domain-generality claim. Bioinformatics is a natural first extension: multi-step pipelines (e.g., read quality control (QC) $\rightarrow$ alignment $\rightarrow$ variant calling $\rightarrow$ annotation) are already expressed as validated DAGs in platforms such as nf-core~\cite{Ewels2020}, and the schema-gated pattern could wrap these pipelines with conversational parameter completion and cross-step type checking while preserving their existing provenance infrastructure. Food science and semiconductor processing offer analogous opportunities where established tool chains could be schema-wrapped without reimplementation.
  \item \emph{AI-assisted tool and workflow authoring}: if an LLM can draft conformant schemas or compose novel workflow DAGs on demand---subject to the same validation and review gates---registry coverage becomes a soft rather than hard constraint, potentially lifting the \cf ceiling imposed by schema-gating.
  \item \emph{Hybrid execution modes}: whether sandboxed generative ``escape hatches'' can coexist with schema-gated guarantees.
  \item \emph{Rubric validation for self-assessment}: evaluating whether the \ed/\cf{} rubric can serve as a reliable self-assessment instrument for practitioners and platform vendors to locate their own systems in the design space and inform architectural roadmaps.
\end{enumerate}

% ---------- Acknowledgments / Contributions / Competing interests ----------
\section*{Acknowledgments}
We thank the research teams and R\&D organisations who participated in interviews and provided feedback that informed the requirements and design considerations in this work. We also thank colleagues at Intellegens for discussions and support during the development of this work.

\section*{Contributions}
J.S. (Joel Strickland) conceived and led the study, developed the survey framework and interview protocol, conducted interviews, synthesised requirements, designed the multi-model LLM scoring methodology and cross-model agreement analysis, assessed and scored all 20 software systems, designed the reference architecture, implemented architectural prototypes, performed the thematic and statistical analyses, and wrote the manuscript.
A.V. (Arjun Vijeta) supported interviews, contributed to synthesis of requirements, contributed to architecture discussions, and provided engineering and implementation support.
C.M. (Chris Moores) provided engineering and implementation support and best-practice input.
O.B. (Oliwia Bodek) provided writing support, editing, and conceptual feedback.
B.N. (Bogdan Nenchev), T.W. (Thomas Whitehead), C.P. (Charles Phillips), K.T. (Karl Tassenberg), and G.C. (Gareth Conduit) reviewed the manuscript, contributed comments and suggested improvements, and provided domain and technical feedback.
B.P. (Ben Pellegrini) secured project funding and provided supervision and oversight.
All authors reviewed and approved the final manuscript.

\section*{Competing interests}
J.S., A.V., B.N., T.W., C.P., K.T., G.C., and B.P. are affiliated with Intellegens, which develops and commercialises the Alchemite\tm\ platform~\cite{Conduit2017}. Interview participants were recruited through Intellegens' customer relationships (\S\ref{sec:interviews}); the Alchemite\tm\ platform is not the system proposed in this work. O.B. and C.M. are independent researchers. The authors declare no other competing interests.

% ---------- Appendix A ----------
\clearpage
\FloatBarrier
\appendix
% Reset table/figure counters and prefix with appendix letter (e.g. Table A.1)
\setcounter{table}{0}
\renewcommand{\thetable}{\Alph{section}.\arabic{table}}
\setcounter{figure}{0}
\renewcommand{\thefigure}{\Alph{section}.\arabic{figure}}
\counterwithin{table}{section}
\counterwithin{figure}{section}
\section{Interview Methods}
\label{app:interview-methods}

\subsection{Recruitment, confidentiality, and study context}
Participants were recruited through Intellegens' professional network (Table~\ref{tab:participant-demographics}). Interviews concerned workflow challenges and expectations for conversational AI in R\&D broadly, not any specific commercial platform. All participants were covered by non-disclosure agreements; we report findings only in aggregated form without attribution. Participants were invited after we explained the study aim and participated voluntarily.

\paragraph{Recruitment scope and limitations.}
All stakeholders were recruited through Intellegens' existing customer relationships, and several authors are Intellegens employees (\S\,Competing Interests). This convenience sample enabled access to experienced R\&D practitioners but limits generalisability: it may over-represent organisations already investing in ML-assisted workflows and under-represent those relying on generative or ad~hoc tooling. No Intellegens product is the system proposed here, and questions were framed around general workflow challenges. Independent replication with broader recruitment would strengthen external validity (\S\ref{sec:future-work}).

Interviewees knew that multiple stakeholders were participating but were not informed of each other's identities. All received the same structure and question set. No personal data beyond professional role/context was collected.

\begin{table}[H]
  \centering
  \small
  \caption{Anonymised participant demographics. Revenue and employee figures are approximate bands based on publicly available data (fiscal year 2025 or latest available). The independent consultant's career spans multiple specialty-chemicals firms. Pseudonym assignment is randomised.}
  \label{tab:participant-demographics}
  \setlength{\tabcolsep}{4pt}
  \renewcommand{\arraystretch}{1.5}
  \begin{tabularx}{\linewidth}{@{}l X r r c@{}}
    \toprule
    \textbf{ID} & \textbf{Industry sector} & \textbf{Revenue (USD)} & \textbf{Employees} & \textbf{Experts} \\
    \midrule
    \rowcolor{gray!6}
    Org-1 & Food \& beverage            & $>$\$50B   & $>$100{,}000   & 1 \\
    Org-2 & Professional tools           & \$5--10B   & 10{,}000+      & 2 \\
    \rowcolor{gray!6}
    Org-3 & Packaging \& materials       & \$5--10B   & 10{,}000--50{,}000 & 2 \\
    Org-4 & Food ingredients             & \$5--10B   & 10{,}000--15{,}000 & 6 \\
    \rowcolor{gray!6}
    Org-5 & Consumer products            & \$5--10B   & 5{,}000--10{,}000  & 1 \\
    Org-6 & Specialty chemicals          & \$1--5B    & 5{,}000--10{,}000  & 2 \\
    \rowcolor{gray!6}
    Org-7 & Semiconductor materials      & \$1--5B    & 5{,}000--10{,}000  & 1 \\
    Org-8 & Construction chemicals       & $\sim$\$1B & 1{,}000--5{,}000   & 1 \\
    \rowcolor{gray!6}
    Org-9 & Industrial printing          & $<$\$1B    & 1{,}000--5{,}000   & 1 \\
    \addlinespace[8pt]
    Consultant & Specialty chemicals (ind.)  & ---        & ---              & 1 \\
    \midrule
    \multicolumn{4}{@{}l}{\textit{Total}} & \textit{18} \\
    \bottomrule
  \end{tabularx}
\end{table}

\subsection{Interview structure}
We conducted up to two sessions per stakeholder:
\begin{itemize}
  \item \textbf{Session 1: Workflow challenges and automation opportunities.} Focused on current computational workflows and friction points: time sinks, manual steps, integration pain points, and situations where the cost of trying alternatives prevents exploration. Questions covered data preparation, model building, analysis, reporting, and handoffs. We also probed where automation---from guided assistance to autonomous execution---could accelerate work.
  \item \textbf{Session 2: Expectations for agentic AI in R\&D.} Examined expectations for conversational and agentic AI: desired autonomy levels, trust and control boundaries, governance requirements, preferred interaction styles (stepwise vs.\ autonomous), transparency needs, deployment constraints (cloud/hybrid/on-prem), and required integrations with existing systems.
\end{itemize}
Interviews followed a designated guide and question schedule while allowing follow-up questions for clarification.

\subsection{Analysis approach}
Themes were coded from raw transcripts (not intermediate summaries). Transcripts were parsed into speaker-turn paragraphs; interviewer and internal-team utterances were excluded, leaving 2{,}468~external-participant paragraphs as coding units.

\subsection{Coding protocol and codebook}
\label{app:codebook}
Each external participant's utterance constitutes one coding unit (2{,}468~paragraphs total). A codebook of 17~themes was developed through iterative reading of all transcripts before systematic coding (Table~\ref{tab:codebook}). Each theme is defined by a label, scope description, and keyword patterns.

Coding used regex-based pattern matching applied uniformly to all paragraphs, with 20~patterns per theme (see \S\ref{sec:thematic-landscape} for rationale). A paragraph was coded for a theme if any associated pattern matched; multiple themes could apply. The 1{,}135~total codings were aggregated by stakeholder breadth and mention frequency (Table~\ref{tab:thematic-groups}).

\paragraph{Methodological note.}
Our approach is systematic content coding, not the iterative interpretive process of reflexive thematic analysis~\cite{BraunClarke2006}. Regex-based matching ensures reproducibility but introduces a precision--recall trade-off: sparse themes may under-count conceptually relevant utterances, while rich themes may over-count incidental mentions. Fixing the count at 20~patterns per theme trades precision for comparability across themes.

To quantify this trade-off, we conducted a three-part sensitivity analysis (full results in Appendix~\ref{app:sensitivity}):
\begin{enumerate}[nosep,leftmargin=*]
  \item \emph{Leave-one-out pattern stability.} Mean maximum single-pattern drop across themes is 32.8\%, with highest dependence in Security \& IP (64.2\%) and Progressive Autonomy (52.6\%)---conceptually narrow themes. Mean per-pattern drop is only 4.1\%, indicating adequate redundancy.
  \item \emph{Cross-meeting convergent validity.} Comparing the two interview rounds ($n_1 = 1{,}023$, $n_2 = 1{,}445$~paragraphs) yields $\rho = 0.487$ ($p = 0.047$), confirming theme salience across sessions.
  \item \emph{Bootstrap confidence intervals.} Resampling (1{,}000~iterations) produces narrow 95\%~CIs (e.g., System Integration: 191, CI~$[166,\,218]$; Progressive Autonomy: 19, CI~$[11,\,28]$), confirming stable frequency estimates.
\end{enumerate}
A spot-check (10~paragraphs per theme) yielded 86--93\% precision; the main false-positive source was polysemous patterns (e.g., \texttt{reason(ing)?} matching ``a reason to believe''). The theme \emph{rankings} motivating architectural categories are robust; absolute counts should be treated as indicative estimates.

\begin{table}[H]
  \centering
  \small
  \caption{Codebook: theme definitions used for coding practitioner interview responses. Row colour encodes thematic group: determinism, flexibility, cross-cutting, or out of scope (see Table~\ref{tab:thematic-groups}).}
  \label{tab:codebook}
  \setlength{\tabcolsep}{8pt}
  \renewcommand{\arraystretch}{1.7}
  \hyphenpenalty=10000
  \exhyphenpenalty=10000
  \begin{tabularx}{\linewidth}{@{}>{\raggedright}p{4.5cm} X@{}}
    \toprule
    \textbf{Theme} & \textbf{Description / Coding Scope} \\
    \midrule
    \rowcolor{thDeterminism!10}
    \emph{System Integration} & Integration with specific named systems: LIMS, ELN, SQL, APIs, etc. \\
    \rowcolor{thDeterminism!10}
    \emph{Workflow \& Task Automation} & Automating repetitive R\&D tasks: design of experiments (DOE), literature review, report generation, experiment design \\
    \rowcolor{thDeterminism!10}
    \emph{Data Cleaning \& Preparation} & Automated data cleaning, formatting, standardisation, quality checks, outlier removal \\
    \rowcolor{thDeterminism!10}
    \emph{Security \& IP Protection} & Data security, IP protection, confidentiality, compliance, access controls \\
    \rowcolor{thDeterminism!10}
    \emph{Domain / Scientific Expertise} & AI understanding domain-specific context, scientific knowledge, industry standards \\
    \rowcolor{thDeterminism!10}
    \emph{Safety \& Ethical Boundaries} & Tasks that should never be automated: safety, ethics, strategic decisions, compliance \\
    \addlinespace[8pt]
    \rowcolor{thFlexibility!10}
    \emph{Data Search \& Retrieval} & Cross-source search, historical data access, knowledge discovery, retrieving information \\
    \rowcolor{thFlexibility!10}
    \emph{Natural Language Interface} & Conversational AI interaction, NLP queries, chatbot-like interaction \\
    \rowcolor{thFlexibility!10}
    \emph{Visualisation \& Reporting} & Charts, plots, PowerPoint/Word outputs, visual communication of results \\
    \rowcolor{thFlexibility!10}
    \emph{Multi-Agent / Agent-to-Agent} & Multiple agents working together, agent-to-agent communication, orchestration \\
    \addlinespace[8pt]
    \rowcolor{thCrossCut!10}
    \emph{Explainability \& Transparency} & Explainable AI reasoning, showing data sources, justifying recommendations, visibility into steps \\
    \rowcolor{thCrossCut!10}
    \emph{Human Oversight \& Control} & Human-in-the-loop, approve/revert steps, override AI, maintain decision authority \\
    \rowcolor{thCrossCut!10}
    \emph{Progressive Autonomy} & Start with high control then gradually increase autonomy as trust builds \\
    \addlinespace[8pt]
    \rowcolor{thOutScope!10}
    \emph{Cost \& Efficiency} & ROI, cost optimisation, resource efficiency, time savings, accelerating R\&D \\
    \rowcolor{thOutScope!10}
    \emph{Adoption \& Change Management} & User adoption, training, cultural barriers, scepticism, learning curve \\
    \rowcolor{thOutScope!10}
    \emph{Deployment Model Preferences} & On-prem, private cloud, hybrid, cloud-hosted, desktop deployment preferences \\
    \rowcolor{thOutScope!10}
    \emph{Data Aggregation \& Anonymisation} & Aggregating and anonymising sensitive customer data for shared use \\
    \bottomrule
  \end{tabularx}
\end{table}

% ────────────────────────────────────────────────────────
% Appendix B: Sensitivity Analysis
% ────────────────────────────────────────────────────────
\section{Thematic Coding Sensitivity Analysis}
\label{app:sensitivity}

This appendix reports the full three-part sensitivity analysis summarised in \S\ref{app:codebook}, operating on the same 2{,}468~paragraphs and 17$\times$20 regex codebook. Code and data are available in the supplementary materials.

\subsection{Pattern contribution and leave-one-out stability}

Of the 340~total regex patterns (20~per theme), 191~(56\%) matched at least one paragraph (mean 11.2~active patterns per theme). Table~\ref{tab:loo-stability} reports, for each theme, the full mention count, the maximum percentage drop caused by removing a single pattern, and the identity of that most-impactful pattern.

\begin{table}[H]
  \centering
  \small
  \caption{Leave-one-out pattern stability. \emph{Max drop} is the largest percentage decrease in coded paragraphs caused by removing any single pattern from the theme's 20-pattern set.}
  \label{tab:loo-stability}
  \setlength{\tabcolsep}{4pt}
  \begin{tabular}{@{}lrrrl@{}}
    \toprule
    \textbf{Theme} & \textbf{Count} & \textbf{Max drop} & \textbf{Mean drop} & \textbf{Most impactful pattern} \\
    \midrule
    System Integration       & 191 & 23.0\% & 3.9\% & \texttt{API[s]?} \\
    Workflow Automation       & 112 & 29.5\% & 4.1\% & \texttt{DOE} \\
    Data Search \& Retrieval  & 108 & 17.6\% & 4.2\% & \texttt{quer(y|ying|ies)} \\
    Natural Language Interface& 86  & 23.3\% & 4.7\% & \texttt{prompt(s|ing)?} \\
    Explainability            & 83  & 30.1\% & 4.5\% & \texttt{explain(ability|able)?} \\
    Data Cleaning             & 70  & 32.9\% & 4.4\% & \texttt{data\textbackslash s+prep} \\
    Security \& IP            & 67  & 64.2\% & 4.1\% & \texttt{secur(e|ity)} \\
    Visualisation             & 65  & 21.5\% & 3.9\% & \texttt{plot(s|ting)?} \\
    Domain Expertise          & 62  & 51.6\% & 4.7\% & \texttt{formulat(ion|e|ing)} \\
    Cost \& Efficiency        & 58  & 19.0\% & 4.6\% & \texttt{time\textbackslash s+(saving|spent)} \\
    Adoption                  & 51  & 33.3\% & 4.1\% & \texttt{adoption} \\
    Human Oversight           & 44  & 29.5\% & 3.4\% & \texttt{human.in.the.loop} \\
    Deployment Preferences    & 43  & 23.3\% & 3.5\% & \texttt{cloud.?host} \\
    Multi-Agent               & 41  & 39.0\% & 4.8\% & \texttt{agentic.*orchestrat} \\
    Safety \& Ethics          & 25  & 28.0\% & 4.0\% & \texttt{never\textbackslash s+automat} \\
    Progressive Autonomy      & 19  & 52.6\% & 4.0\% & \texttt{(build|grow).*trust} \\
    Data Anonymisation        & 10  & 40.0\% & 4.0\% & \texttt{aggregat.*data} \\
    \bottomrule
  \end{tabular}
\end{table}

Three themes---Security \& IP (64.2\%), Progressive Autonomy (52.6\%), and Domain Expertise (51.6\%)---exhibit high single-pattern concentration. These are conceptually narrow themes whose vocabularies cluster around core keywords (\emph{security}, \emph{trust-building}, \emph{formulation}). The remaining 14~themes have maximum drops below 40\%, and in all cases the mean per-pattern drop is $\leq$5\%, indicating adequate redundancy across the pattern set.

\subsection{Cross-meeting convergent validity}

The two sessions ($n_1 = 1{,}023$, $n_2 = 1{,}445$~paragraphs) yield $\rho = 0.487$ ($p = 0.047$), confirming consistent theme salience despite different structures and deeper Session~2 probing (which increased System Integration from 49 to 142~mentions and Security from 12 to 55). Themes appearing predominantly in one session (e.g., Deployment Preferences: 1~vs.~42; Multi-Agent: 2~vs.~39) reflect Session~2's agentic-AI focus.

\subsection{Bootstrap confidence intervals}

Resampling 2{,}468~paragraphs with replacement (1{,}000~iterations) produces the 95\% confidence intervals in Table~\ref{tab:bootstrap-ci}.

\begin{table}[H]
  \centering
  \small
  \caption{Bootstrap 95\% confidence intervals for theme mention counts (1{,}000~iterations).}
  \label{tab:bootstrap-ci}
  \setlength{\tabcolsep}{6pt}
  \begin{tabular}{@{}lrr@{}}
    \toprule
    \textbf{Theme} & \textbf{Observed} & \textbf{95\% CI} \\
    \midrule
    System Integration       & 191 & [166, 218] \\
    Workflow Automation       & 112 & [94, 132] \\
    Data Search \& Retrieval  & 108 & [88, 128] \\
    Natural Language Interface& 86  & [68, 104] \\
    Explainability            & 83  & [66, 101] \\
    Data Cleaning             & 70  & [54, 88] \\
    Security \& IP            & 67  & [51, 84] \\
    Visualisation             & 65  & [50, 80] \\
    Domain Expertise          & 62  & [47, 78] \\
    Cost \& Efficiency        & 58  & [44, 74] \\
    Adoption                  & 51  & [37, 65] \\
    Human Oversight           & 44  & [33, 57] \\
    Deployment Preferences    & 43  & [31, 55] \\
    Multi-Agent               & 41  & [29, 54] \\
    Safety \& Ethics          & 25  & [16, 35] \\
    Progressive Autonomy      & 19  & [11, 28] \\
    Data Anonymisation        & 10  & [5, 17] \\
    \bottomrule
  \end{tabular}
\end{table}

Median CI half-width is $\approx$15\% of observed values. The top-five ranking is preserved across all 1{,}000~samples, confirming that architectural conclusions are not artefacts of particular paragraphs or companies.

% ---------- Appendix C: Scoring Protocol ----------
\section{Scoring Protocol}
\label{app:scoring-protocol}

The following prompt was submitted to ChatGPT~5.2 (OpenAI, 2025), Claude Sonnet~4.6 (Anthropic, 2025; via Perplexity), and Gemini~3.1~Pro (Google, 2025; via Perplexity)---all with thinking mode enabled---once per system in a fresh context with no prior scoring history (five sessions per model, 15~total). The rubric is identical to Table~\ref{tab:table2}.

\begin{quote}
\small
\textbf{System Rules}\\
You are scoring a software system on two architectural dimensions using the rubric below. Score based on default, built-in capabilities for the system as publicly documented. Do not score based on hypothetical configurations or custom extensions. Use web search to verify capabilities using official documentation, product pages, academic papers, or reputable technical sources. Prefer primary sources. When uncertain, choose the lower score and explain your reasoning. Every score must be justified by objective, observable evidence or clear reasoning for inference.

\medskip
\textbf{System Being Scored}\\
Name: [System Name]\\
Website / Documentation: [URL(s)]\\
Version / Tier: [as publicly documented]

\medskip
\textbf{Scoring Rubric}\\
Score each dimension on a 1--5 ordinal scale:

\medskip
\emph{Execution Determinism (ED):}\\
1 = Unconstrained generated code; no formal pre-execution gate\\
2 = Primarily runtime checks; weak pre-execution constraints\\
3 = Developer-defined tool interfaces; partial validation but not uniformly schema-governed\\
4 = Schema/type-validated tool calls with strong pre-execution validation\\
5 = Explicit workflow spec/registry with static validation; reproducibility by construction

\medskip
\emph{Conversational Flexibility (CF):}\\
1 = No conversational interface (DSL/config/GUI only)\\
2 = Limited NL assistance (form help, search, hints); NL does not control execution\\
3 = NL assists authoring/modifying workflows; execution requires an externalized artifact\\
4 = NL directly selects/parameterizes tool/workflow invocations via structured calls\\
5 = Free-form NL or agentic loops directly driving actions with minimal structural constraint

\medskip
\textbf{Required Output}\\
For each dimension provide: (1)~score (1--5), (2)~confidence (high/medium/low), (3)~1--2 sentence justification citing specific evidence, and (4)~boundary-case notes if applicable.
\end{quote}

\subsection{Per-session scores}
\label{app:per-session-scores}
Tables~\ref{tab:per-session-chatgpt}--\ref{tab:per-session-gemini} report the individual \ed/\cf{} scores from each of the 15~independent scoring sessions described in \S\ref{sec:ordinal-scoring}, grouped by model family. Each session used the prompt template above in a fresh conversational context. Row colours encode paradigm groups (see Table~\ref{tab:table1}).

\begin{table}[H]
  \centering
  \small
  \caption{Per-session \ed/\cf{} scores: ChatGPT~5.2 (Runs~1--5). Each cell shows (\ed,\,\cf). Med.\ = within-model median. Row colours encode the paradigm groups defined in Table~\ref{tab:table1}.}
  \label{tab:per-session-chatgpt}
  \setlength{\tabcolsep}{3pt}
  \renewcommand{\arraystretch}{1.4}
  \adjustbox{max width=\textwidth}{%
  \begin{tabular}{@{}c l c c c c c c c@{}}
    \toprule
    \textbf{ID} & \textbf{System} & \textbf{Run 1} & \textbf{Run 2} & \textbf{Run 3} & \textbf{Run 4} & \textbf{Run 5} & \textbf{Med.\ \ed} & \textbf{Med.\ \cf} \\
    \midrule
    \rowcolor{grpGenerative!10}
     1 & LLM chat (generic) & (1,\,3) & (1,\,3) & (1,\,3) & (1,\,3) & (1,\,3) & 1 & 3 \\
    \rowcolor{grpGenerative!10}
     2 & GitHub Copilot & (2,\,5) & (3,\,5) & (3,\,5) & (2,\,5) & (2,\,5) & 2 & 5 \\
    \rowcolor{grpGenerative!10}
     3 & AutoGPT-style agents & (2,\,5) & (2,\,5) & (2,\,5) & (3,\,5) & (2,\,5) & 2 & 5 \\
    \rowcolor{grpGenerative!10}
     4 & ReAct-style fwks & (3,\,5) & (3,\,5) & (3,\,5) & (3,\,5) & (3,\,5) & 3 & 5 \\
    \addlinespace[10pt]
    \rowcolor{grpToolAug!10}
     5 & LangChain (typical) & (3,\,5) & (3,\,5) & (3,\,5) & (3,\,5) & (3,\,5) & 3 & 5 \\
    \rowcolor{grpToolAug!10}
     6 & SciAgents & (3,\,5) & (3,\,5) & (3,\,5) & (2,\,5) & (3,\,5) & 3 & 5 \\
    \rowcolor{grpToolAug!10}
     7 & Semantic Kernel & (3,\,4) & (3,\,4) & (3,\,4) & (3,\,4) & (3,\,4) & 3 & 4 \\
    \addlinespace[10pt]
    \rowcolor{grpAgentic!10}
     8 & OpenAI Assistants & (4,\,4) & (4,\,4) & (4,\,4) & (4,\,4) & (4,\,4) & 4 & 4 \\
    \rowcolor{grpAgentic!10}
     9 & Copilot Studio & (4,\,4) & (4,\,4) & (5,\,4) & (4,\,4) & (4,\,4) & 4 & 4 \\
    \addlinespace[10pt]
    \rowcolor{grpWfNL!10}
    10 & n8n & (3,\,3) & (3,\,3) & (3,\,3) & (3,\,3) & (3,\,3) & 3 & 3 \\
    \rowcolor{grpWfNL!10}
    11 & Snakemaker-style & (3,\,3) & (3,\,3) & (4,\,3) & (3,\,3) & (3,\,3) & 3 & 3 \\
    \rowcolor{grpWfNL!10}
    12 & Dataiku DSS & (5,\,3) & (3,\,3) & (4,\,3) & (4,\,3) & (4,\,3) & 4 & 3 \\
    \addlinespace[10pt]
    \rowcolor{grpWfCentric!10}
    13 & Galaxy & (5,\,1) & (5,\,1) & (5,\,2) & (5,\,1) & (5,\,2) & 5 & 1 \\
    \rowcolor{grpWfCentric!10}
    14 & Snakemake & (5,\,1) & (3,\,1) & (5,\,1) & (3,\,1) & (5,\,1) & 5 & 1 \\
    \rowcolor{grpWfCentric!10}
    15 & Nextflow & (3,\,1) & (3,\,1) & (3,\,1) & (3,\,1) & (3,\,1) & 3 & 1 \\
    \rowcolor{grpWfCentric!10}
    16 & nf-core & (4,\,1) & (4,\,1) & (4,\,1) & (4,\,1) & (4,\,1) & 4 & 1 \\
    \rowcolor{grpWfCentric!10}
    17 & WorkflowHub-style & (4,\,1) & (4,\,1) & (4,\,1) & (4,\,1) & (4,\,1) & 4 & 1 \\
    \rowcolor{grpWfCentric!10}
    18 & AWS Step Functions & (5,\,2) & (5,\,1) & (5,\,1) & (5,\,1) & (5,\,1) & 5 & 1 \\
    \rowcolor{grpWfCentric!10}
    19 & Apache Airflow & (3,\,1) & (3,\,1) & (3,\,1) & (3,\,1) & (3,\,1) & 3 & 1 \\
    \rowcolor{grpWfCentric!10}
    20 & Kubeflow / Argo & (5,\,1) & (5,\,1) & (5,\,1) & (5,\,1) & (5,\,1) & 5 & 1 \\
    \bottomrule
  \end{tabular}%
  }
\end{table}

\begin{table}[H]
  \centering
  \small
  \caption{Per-session \ed/\cf{} scores: Claude Sonnet~4.6 (via Perplexity) (Runs~1--5). Each cell shows (\ed,\,\cf). Med.\ = within-model median. Row colours encode the paradigm groups defined in Table~\ref{tab:table1}.}
  \label{tab:per-session-claude}
  \setlength{\tabcolsep}{3pt}
  \renewcommand{\arraystretch}{1.4}
  \adjustbox{max width=\textwidth}{%
  \begin{tabular}{@{}c l c c c c c c c@{}}
    \toprule
    \textbf{ID} & \textbf{System} & \textbf{Run 1} & \textbf{Run 2} & \textbf{Run 3} & \textbf{Run 4} & \textbf{Run 5} & \textbf{Med.\ \ed} & \textbf{Med.\ \cf} \\
    \midrule
    \rowcolor{grpGenerative!10}
     1 & LLM chat (generic) & (1,\,3) & (1,\,3) & (1,\,3) & (1,\,3) & (1,\,3) & 1 & 3 \\
    \rowcolor{grpGenerative!10}
     2 & GitHub Copilot & (3,\,5) & (2,\,5) & (2,\,5) & (3,\,5) & (3,\,5) & 3 & 5 \\
    \rowcolor{grpGenerative!10}
     3 & AutoGPT-style agents & (2,\,5) & (2,\,5) & (2,\,5) & (2,\,5) & (2,\,5) & 2 & 5 \\
    \rowcolor{grpGenerative!10}
     4 & ReAct-style fwks & (2,\,5) & (2,\,5) & (2,\,5) & (2,\,5) & (2,\,5) & 2 & 5 \\
    \addlinespace[10pt]
    \rowcolor{grpToolAug!10}
     5 & LangChain (typical) & (3,\,5) & (3,\,5) & (3,\,5) & (3,\,5) & (3,\,5) & 3 & 5 \\
    \rowcolor{grpToolAug!10}
     6 & SciAgents & (3,\,5) & (2,\,5) & (2,\,5) & (3,\,5) & (2,\,5) & 2 & 5 \\
    \rowcolor{grpToolAug!10}
     7 & Semantic Kernel & (4,\,4) & (3,\,4) & (3,\,4) & (3,\,4) & (3,\,4) & 3 & 4 \\
    \addlinespace[10pt]
    \rowcolor{grpAgentic!10}
     8 & OpenAI Assistants & (4,\,4) & (4,\,4) & (4,\,4) & (4,\,4) & (4,\,4) & 4 & 4 \\
    \rowcolor{grpAgentic!10}
     9 & Copilot Studio & (4,\,4) & (5,\,4) & (4,\,4) & (4,\,3) & (5,\,4) & 4 & 4 \\
    \addlinespace[10pt]
    \rowcolor{grpWfNL!10}
    10 & n8n & (4,\,3) & (4,\,3) & (4,\,3) & (4,\,3) & (4,\,3) & 4 & 3 \\
    \rowcolor{grpWfNL!10}
    11 & Snakemaker-style & (5,\,3) & (5,\,3) & (5,\,3) & (5,\,3) & (5,\,3) & 5 & 3 \\
    \rowcolor{grpWfNL!10}
    12 & Dataiku DSS & (4,\,3) & (4,\,3) & (4,\,3) & (4,\,3) & (4,\,3) & 4 & 3 \\
    \addlinespace[10pt]
    \rowcolor{grpWfCentric!10}
    13 & Galaxy & (5,\,2) & (5,\,2) & (5,\,2) & (5,\,2) & (5,\,2) & 5 & 2 \\
    \rowcolor{grpWfCentric!10}
    14 & Snakemake & (5,\,1) & (5,\,1) & (5,\,1) & (5,\,1) & (5,\,1) & 5 & 1 \\
    \rowcolor{grpWfCentric!10}
    15 & Nextflow & (5,\,1) & (4,\,1) & (5,\,1) & (5,\,1) & (5,\,1) & 5 & 1 \\
    \rowcolor{grpWfCentric!10}
    16 & nf-core & (5,\,1) & (5,\,1) & (5,\,1) & (5,\,1) & (5,\,1) & 5 & 1 \\
    \rowcolor{grpWfCentric!10}
    17 & WorkflowHub-style & (5,\,1) & (5,\,1) & (5,\,1) & (5,\,1) & (5,\,1) & 5 & 1 \\
    \rowcolor{grpWfCentric!10}
    18 & AWS Step Functions & (5,\,1) & (5,\,2) & (5,\,1) & (5,\,1) & (5,\,1) & 5 & 1 \\
    \rowcolor{grpWfCentric!10}
    19 & Apache Airflow & (5,\,1) & (5,\,1) & (4,\,1) & (5,\,1) & (5,\,1) & 5 & 1 \\
    \rowcolor{grpWfCentric!10}
    20 & Kubeflow / Argo & (5,\,1) & (5,\,1) & (5,\,1) & (5,\,1) & (5,\,1) & 5 & 1 \\
    \bottomrule
  \end{tabular}%
  }
\end{table}

\begin{table}[H]
  \centering
  \small
  \caption{Per-session \ed/\cf{} scores: Gemini~3.1~Pro (via Perplexity) (Runs~1--5). Each cell shows (\ed,\,\cf). Med.\ = within-model median. Row colours encode the paradigm groups defined in Table~\ref{tab:table1}.}
  \label{tab:per-session-gemini}
  \setlength{\tabcolsep}{3pt}
  \renewcommand{\arraystretch}{1.4}
  \adjustbox{max width=\textwidth}{%
  \begin{tabular}{@{}c l c c c c c c c@{}}
    \toprule
    \textbf{ID} & \textbf{System} & \textbf{Run 1} & \textbf{Run 2} & \textbf{Run 3} & \textbf{Run 4} & \textbf{Run 5} & \textbf{Med.\ \ed} & \textbf{Med.\ \cf} \\
    \midrule
    \rowcolor{grpGenerative!10}
     1 & LLM chat (generic) & (1,\,3) & (1,\,3) & (1,\,3) & (1,\,5) & (1,\,3) & 1 & 3 \\
    \rowcolor{grpGenerative!10}
     2 & GitHub Copilot & (1,\,5) & (1,\,5) & (3,\,5) & (1,\,5) & (2,\,5) & 1 & 5 \\
    \rowcolor{grpGenerative!10}
     3 & AutoGPT-style agents & (1,\,5) & (2,\,5) & (1,\,5) & (2,\,5) & (2,\,5) & 2 & 5 \\
    \rowcolor{grpGenerative!10}
     4 & ReAct-style fwks & (2,\,5) & (2,\,5) & (2,\,5) & (2,\,5) & (2,\,5) & 2 & 5 \\
    \addlinespace[10pt]
    \rowcolor{grpToolAug!10}
     5 & LangChain (typical) & (4,\,5) & (4,\,5) & (4,\,5) & (4,\,5) & (3,\,5) & 4 & 5 \\
    \rowcolor{grpToolAug!10}
     6 & SciAgents & (3,\,5) & (1,\,5) & (3,\,5) & (2,\,5) & (3,\,5) & 3 & 5 \\
    \rowcolor{grpToolAug!10}
     7 & Semantic Kernel & (3,\,4) & (3,\,4) & (3,\,5) & (3,\,4) & (3,\,4) & 3 & 4 \\
    \addlinespace[10pt]
    \rowcolor{grpAgentic!10}
     8 & OpenAI Assistants & (4,\,4) & (4,\,4) & (4,\,4) & (4,\,4) & (4,\,4) & 4 & 4 \\
    \rowcolor{grpAgentic!10}
     9 & Copilot Studio & (5,\,4) & (5,\,4) & (4,\,4) & (4,\,4) & (5,\,4) & 5 & 4 \\
    \addlinespace[10pt]
    \rowcolor{grpWfNL!10}
    10 & n8n & (4,\,3) & (4,\,3) & (4,\,3) & (5,\,3) & (3,\,3) & 4 & 3 \\
    \rowcolor{grpWfNL!10}
    11 & Snakemaker-style & (5,\,3) & (5,\,3) & (5,\,3) & (5,\,3) & (5,\,3) & 5 & 3 \\
    \rowcolor{grpWfNL!10}
    12 & Dataiku DSS & (5,\,3) & (4,\,3) & (5,\,3) & (5,\,3) & (5,\,3) & 5 & 3 \\
    \addlinespace[10pt]
    \rowcolor{grpWfCentric!10}
    13 & Galaxy & (5,\,2) & (5,\,2) & (5,\,1) & (5,\,3) & (5,\,2) & 5 & 2 \\
    \rowcolor{grpWfCentric!10}
    14 & Snakemake & (5,\,1) & (5,\,1) & (5,\,1) & (3,\,1) & (5,\,1) & 5 & 1 \\
    \rowcolor{grpWfCentric!10}
    15 & Nextflow & (3,\,1) & (3,\,1) & (4,\,1) & (3,\,1) & (3,\,1) & 3 & 1 \\
    \rowcolor{grpWfCentric!10}
    16 & nf-core & (5,\,1) & (5,\,1) & (5,\,1) & (5,\,1) & (5,\,1) & 5 & 1 \\
    \rowcolor{grpWfCentric!10}
    17 & WorkflowHub-style & (5,\,1) & (5,\,1) & (5,\,1) & (5,\,1) & (5,\,1) & 5 & 1 \\
    \rowcolor{grpWfCentric!10}
    18 & AWS Step Functions & (5,\,3) & (5,\,1) & (5,\,1) & (5,\,1) & (5,\,1) & 5 & 1 \\
    \rowcolor{grpWfCentric!10}
    19 & Apache Airflow & (5,\,1) & (5,\,1) & (5,\,1) & (4,\,1) & (5,\,1) & 5 & 1 \\
    \rowcolor{grpWfCentric!10}
    20 & Kubeflow / Argo & (5,\,1) & (5,\,1) & (5,\,1) & (5,\,1) & (4,\,1) & 5 & 1 \\
    \bottomrule
  \end{tabular}%
  }
\end{table}

\noindent\textbf{Within-model consistency.}
ChatGPT~5.2: 31/40 score assignments identical across five runs, 38/40 within one ordinal point; Krippendorff's $\alpha_{\mathrm{ED}}=0.875$, $\alpha_{\mathrm{CF}}=0.991$.
Claude Sonnet~4.6: 32/40 identical, 40/40 within one point; $\alpha_{\mathrm{ED}}=0.954$, $\alpha_{\mathrm{CF}}=0.992$.
Gemini~3.1~Pro: 25/40 identical, 33/40 within one point; $\alpha_{\mathrm{ED}}=0.885$, $\alpha_{\mathrm{CF}}=0.957$.

\noindent\textbf{Cross-model agreement.}
Across all 15~runs, 19/40 score assignments were identical and 28/40 varied by at most one ordinal point. Krippendorff's~$\alpha$ (ordinal, treating each run as an independent rater): $\alpha_{\mathrm{ED}}=0.797$, $\alpha_{\mathrm{CF}}=0.980$. Pairwise cross-platform agreement: ChatGPT\,+\,Claude $\alpha_{\mathrm{ED}}=0.788$, $\alpha_{\mathrm{CF}}=0.990$; ChatGPT\,+\,Gemini $\alpha_{\mathrm{ED}}=0.777$, $\alpha_{\mathrm{CF}}=0.973$; Claude\,+\,Gemini $\alpha_{\mathrm{ED}}=0.885$, $\alpha_{\mathrm{CF}}=0.976$. Seven systems saw their consensus median shift by one point relative to the ChatGPT-only median; the overall Pareto-front structure and paradigm clustering are unchanged.

\noindent\textbf{Reflexivity note.}
The ChatGPT scoring model is produced by the same organisation (OpenAI) responsible for one reviewed system (ID~8), introducing a potential reflexivity concern. This is directly mitigated by independent confirmation from two additional model families (Claude, Gemini): ID~8 received identical \ed/\cf{} scores from all three models, consistent with its publicly documented architecture (strong schema validation, partial provenance).

A parallel concern applies to Claude Sonnet~4.6 (Anthropic), whose tool-use API and MCP~\cite{Anthropic2024MCP} feature in the schema-gated discussion. No Anthropic product is separately scored, so the risk is indirect. Claude's scores show no systematic \ed{} bias (pairwise $\alpha_{\mathrm{ED}}$: Claude\,+\,Gemini $= 0.885$, Claude\,+\,ChatGPT $= 0.788$, both substantial agreement), and the Pareto-front structure is unchanged when Claude sessions are excluded.

All consensus median scores were independently confirmed by the authors.

% ---------- Appendix D: Schema Examples ----------
\section{Schema Examples}
\label{app:schema-examples}
This appendix provides the full tool and workflow schemas described in Section~\ref{sec:validation-framework}.

\FloatBarrier
\subsection{Domain Tool Schema Example: Materials-Property Predictor}
\label{app:tool-schema}
The following definition illustrates a domain tool schema. An MCP tool definition would contain the \texttt{parameters} block; the fields below it---\texttt{input\_schema}, \texttt{output\_schema}, \texttt{dependencies}, \texttt{domain\_tags}, \texttt{provenance}, and operational metadata---are the additions enabling composition, discovery, and provenance.

\begin{verbatim}
ToolDefinition(
    id="materials_property_predictor",
    name="Materials-Property Predictor",
    description="Train and evaluate surrogate models for
        composition-property relationships",
    version="2.1.0",
    parameters=[
        ParameterDefinition(
            name="dataset_id",
            type="str",
            description="UUID of uploaded dataset",
            required=True,
            examples=["550e8400-e29b-41d4-a716-446655440000"]
        ),
        ParameterDefinition(
            name="target_properties",
            type="list[str]",
            description="Column names to predict",
            required=True,
            examples=[["yield_strength", "elongation"]]
        ),
        ParameterDefinition(
            name="validation_strategy",
            type="str",
            description="Cross-validation approach",
            required=False,
            default="5-fold",
            allowed_values=["5-fold", "10-fold", "leave-one-out"]
        ),
    ],
    # --- Fields beyond a standard MCP tool definition ---
    input_schema={
        "dataset": {"type": "dataframe", "columns": "dynamic"},
        "target_columns": {"type": "list[string]"}
    },
    output_schema={
        "model_id": {"type": "string"},
        "metrics": {"type": "dict", "keys": ["r2_score", "rmse"]},
        "predictions": {"type": "dataframe"}
    },
    dependencies=["data_loader"],
    domain_tags=["materials", "surrogate-model", "regression"],
    provenance={
        "origin": "intellegens-core",
        "maintainer": "ml-team@example.org"
    },
    estimated_duration=5.0,   # minutes
    requires_network=True
)
\end{verbatim}

Because the schema is a structured format, an LLM can draft a conformant \texttt{ToolDefinition} from existing documentation or code, reducing authoring burden. The draft enters the registry only after passing the same automated validation applied to manual definitions.

\subsection{Workflow Schema Example: Basic Data Analysis}
\label{app:workflow-schema}
The following JSON schema defines a three-step load--clean--analyse workflow. Each step invokes a validated domain tool; data flows explicitly via parameter mappings and graph edges. Workflow-level \texttt{parameters} define user-facing inputs with pre-execution validation rules; \texttt{metadata} supports discovery.

\begin{verbatim}
{
  "workflow_id": "basic_data_analysis",
  "name": "Basic Data Analysis",
  "description": "Load, clean, and statistically profile a
      tabular dataset",
  "steps": [
    {
      "step_id": "load_data",
      "tool_id": "data_loader",
      "name": "Load Data",
      "description": "Load data from various sources",
      "parameters": {"file_type": "csv"},
      "dependencies": [],
      "estimated_duration": 2.0
    },
    {
      "step_id": "clean_data",
      "tool_id": "data_cleaner",
      "name": "Clean Data",
      "description": "Clean and preprocess the data",
      "parameters": {
        "operations": ["remove_duplicates", "handle_missing"],
        "missing_strategy": "remove"
      },
      "dependencies": ["load_data"],
      "estimated_duration": 3.0
    },
    {
      "step_id": "analyze_data",
      "tool_id": "data_analyzer",
      "name": "Analyze Data",
      "description": "Perform statistical analysis",
      "parameters": {
        "analysis_type": "dataset_profile",
        "generate_summary": true
      },
      "dependencies": ["clean_data"],
      "estimated_duration": 5.0
    }
  ],
  "parameter_mappings": [
    {
      "from_step": "load_data",
      "from_parameter": "data",
      "to_step": "clean_data",
      "to_parameter": "data",
      "description": "Pass loaded data to cleaning step"
    },
    {
      "from_step": "clean_data",
      "from_parameter": "cleaned_data",
      "to_step": "analyze_data",
      "to_parameter": "data",
      "description": "Pass cleaned data to analysis step"
    }
  ],
  "edges": [
    {
      "edge_id": "load_to_clean",
      "source_node_id": "load_data",
      "target_node_id": "clean_data",
      "source_output": "data",
      "target_input": "data"
    },
    {
      "edge_id": "clean_to_analyze",
      "source_node_id": "clean_data",
      "target_node_id": "analyze_data",
      "source_output": "cleaned_data",
      "target_input": "data"
    }
  ],
  "parameters": {
    "dataset_file": {
      "type": "string",
      "required": true,
      "description": "Path to the dataset file",
      "validation_rules": {"not_empty": true}
    },
    "missing_strategy": {
      "type": "string",
      "required": false,
      "default": "remove",
      "description": "Strategy for handling missing data",
      "validation_rules": {
        "allowed_values": ["remove", "fill_mean", "fill_median"]
      }
    }
  },
  "metadata": {
    "complexity": "simple",
    "estimated_duration_minutes": 10,
    "tags": ["data", "analysis", "basic"],
    "categories": ["data_analysis"],
    "use_cases": ["Basic data exploration",
                   "Quick data insights"]
  }
}
\end{verbatim}

Together the examples illustrate what distinguishes these schemas from flat tool-calling interfaces: typed contracts enable automatic inter-step wiring; dependency declarations make execution order machine-checkable; validation rules enforce pre-execution constraints; and provenance metadata supports discovery, auditing, and reproducibility.

\FloatBarrier

\bibliographystyle{unsrtnat}
\bibliography{references}

@article{BraunClarke2006,
  author  = {Braun, Virginia and Clarke, Victoria},
  title   = {Using thematic analysis in psychology},
  journal = {Qualitative Research in Psychology},
  volume  = {3},
  number  = {2},
  pages   = {77--101},
  year    = {2006},
  doi     = {10.1191/1478088706qp063oa}
}

@article{Strickland2021,
  author  = {Strickland, J. and Nenchev, B. and Tassenberg, K. and Perry, S. and Sheppard, G. and Dong, H. and Zhang, R. and Burca, G. and D'Souza, N.},
  title   = {On the origin of mosaicity in directionally solidified Ni-base superalloys},
  journal = {Acta Materialia},
  volume  = {217},
  pages   = {117180},
  year    = {2021},
  doi     = {10.1016/j.actamat.2021.117180}
}

@article{CohenBoulakia2017,
  author  = {Cohen-Boulakia, S. and Belhajjame, K. and Collin, O. and Chopard, J. and Froidevaux, C. and Gaignard, A. and Hinsen, K. and Larmande, P. and Le Bras, Y. and Lemoine, F. and Mareuil, F.},
  title   = {Scientific workflows for computational reproducibility in the life sciences: Status, challenges and opportunities},
  journal = {Future Generation Computer Systems},
  volume  = {75},
  pages   = {284--298},
  year    = {2017},
  doi     = {10.1016/j.future.2017.01.012}
}

@article{Hope2023,
  author  = {Hope, T. and Downey, D. and Weld, D. S. and Etzioni, O. and Horvitz, E.},
  title   = {A computational inflection for scientific discovery},
  journal = {Communications of the ACM},
  volume  = {66},
  number  = {8},
  pages   = {62--73},
  year    = {2023},
  doi     = {10.1145/3576896}
}

@misc{Liu2025,
  author       = {Liu, Fan and Han, Jindong and Lyu, Tengfei and Zhang, Weijia and Yang, Zhe-Rui and Dai, Lu and Liu, Cancheng and Liu, Hao},
  title        = {Foundation Models for Scientific Discovery: From Paradigm Enhancement to Paradigm Transition},
  howpublished = {arXiv preprint arXiv:2510.15280},
  year         = {2025},
  doi          = {10.48550/arXiv.2510.15280},
  note         = {Preprint}
}

@article{Huber2021,
  author  = {Huber, S. P. and Bosoni, E. and Bercx, M. and Br{\"o}der, J. and Degomme, A. and Dikan, V. and Eimre, K. and Flage-Larsen, E. and Garcia, A. and Genovese, L. and Gresch, D.},
  title   = {Common workflows for computing material properties using different quantum engines},
  journal = {npj Computational Materials},
  volume  = {7},
  number  = {1},
  pages   = {136},
  year    = {2021},
  doi     = {10.1038/s41524-021-00594-6}
}

@article{Ganose2025,
  author  = {Ganose, A. M. and Sahasrabuddhe, H. and Asta, M. and Beck, K. and Biswas, T. and Bonkowski, A. and Bustamante, J. and Chen, X. and Chiang, Y. and Chrzan, D. C. and Clary, J.},
  title   = {Atomate2: Modular workflows for materials science},
  journal = {Digital Discovery},
  volume  = {4},
  number  = {7},
  pages   = {1944--1973},
  year    = {2025},
  doi     = {10.1039/D5DD00019J}
}

@article{Elton2018,
  author  = {Elton, D. C. and Boukouvalas, Z. and Butrico, M. S. and Fuge, M. D. and Chung, P. W.},
  title   = {Applying machine learning techniques to predict the properties of energetic materials},
  journal = {Scientific Reports},
  volume  = {8},
  number  = {1},
  pages   = {9059},
  year    = {2018},
  doi     = {10.1038/s41598-018-27344-x}
}

@article{Patel2022,
  author  = {Patel, R. A. and Borca, C. H. and Webb, M. A.},
  title   = {Featurization strategies for polymer sequence or composition design by machine learning},
  journal = {Molecular Systems Design \& Engineering},
  volume  = {7},
  number  = {6},
  pages   = {661--676},
  year    = {2022},
  doi     = {10.1039/D1ME00160D}
}

@article{Lamprecht2021,
  author  = {Lamprecht, A. L. and Palmblad, M. and Ison, J. and Schw{\"a}mmle, V. and Al Manir, M. S. and Altintas, I. and Baker, C. J. and Amor, A. B. H. and Capella-Gutierrez, S. and Charonyktakis, P. and Crusoe, M. R.},
  title   = {Perspectives on automated composition of workflows in the life sciences},
  journal = {F1000Research},
  volume  = {10},
  pages   = {897},
  year    = {2021},
  doi     = {10.12688/f1000research.54159.1}
}

@article{Singh2021,
  author  = {Singh, A. and Purawat, S. and Rao, A. and Altintas, I.},
  title   = {Modular performance prediction for scientific workflows using Machine Learning},
  journal = {Future Generation Computer Systems},
  volume  = {114},
  pages   = {1--14},
  year    = {2021},
  doi     = {10.1016/j.future.2020.04.048}
}

@article{Gustafsson2025,
  author  = {Gustafsson, O. J. R. and Wilkinson, S. R. and Bacall, F. and Soiland-Reyes, S. and Leo, S. and Pireddu, L. and Owen, S. and Juty, N. and Fern{\'a}ndez, J. M. and Brown, T. and M{\'e}nager, H.},
  title   = {WorkflowHub: a registry for computational workflows},
  journal = {Scientific Data},
  volume  = {12},
  number  = {1},
  pages   = {837},
  year    = {2025},
  doi     = {10.1038/s41597-025-04786-3}
}

@article{Farshidi2023,
  author  = {Farshidi, S. and Liao, X. and Li, N. and Goldfarb, D. and Magagna, B. and Stocker, M. and Jeffery, K. and Thijsse, P. and Pichot, C. and Petzold, A. and Zhao, Z.},
  title   = {Knowledge sharing and discovery across heterogeneous research infrastructures},
  journal = {Open Research Europe},
  volume  = {1},
  pages   = {68},
  year    = {2023},
  doi     = {10.12688/openreseurope.13677.3}
}

@misc{McInnes2025,
  author       = {McInnes, L. C. and Arnold, D. and Balaprakash, P. and Bernhardt, M. and Cerny, B. and Dubey, A. and Giles, R. and Hood, D. W. and Leung, M. A. and Lopez-Marrero, V. and Messina, P.},
  title        = {Report of the 2025 Workshop on Next-Generation Ecosystems for Scientific Computing: Harnessing Community, Software, and AI for Cross-Disciplinary Team Science},
  howpublished = {arXiv preprint arXiv:2510.03413},
  year         = {2025},
  doi          = {10.48550/arXiv.2510.03413},
  note         = {arXiv}
}

@book{OliveiraStewart2006,
  author    = {Oliveira, S. and Stewart, D. E.},
  title     = {Writing Scientific Software: A Guide to Good Style},
  publisher = {Cambridge University Press},
  year      = {2006},
  isbn      = {9780521675956}
}

@article{Alam2025a,
  author  = {Alam, K. and Roy, B. and Roy, C. K. and Mittal, K.},
  title   = {An empirical investigation on the challenges in scientific workflow systems development},
  journal = {Empirical Software Engineering},
  volume  = {30},
  number  = {5},
  pages   = {151},
  year    = {2025},
  doi     = {10.1007/s10664-025-10705-2}
}

@article{Lee2018,
  author  = {Lee, B. D.},
  title   = {Ten simple rules for documenting scientific software},
  journal = {PLOS Computational Biology},
  volume  = {14},
  number  = {12},
  pages   = {e1006561},
  year    = {2018},
  doi     = {10.1371/journal.pcbi.1006561}
}

@inproceedings{Wang2023a,
  author    = {Wang, J. and Xiao, G. and Zhang, S. and Lei, H. and Liu, Y. and Sui, Y.},
  title     = {Compatibility issues in deep learning systems: Problems and opportunities},
  booktitle = {Proceedings of the 31st ACM Joint European Software Engineering Conference and Symposium on the Foundations of Software Engineering (ESEC/FSE '23)},
  pages     = {476--488},
  year      = {2023},
  month     = {nov},
  doi       = {10.1145/3611643.3616321}
}

@inproceedings{Pimentel2008,
  author    = {Pimentel, A. D. and Stefanov, T. and Nikolov, H. and Thompson, M. and Polstra, S. and Deprettere, E. F.},
  title     = {Tool integration and interoperability challenges of a system-level design flow: A case study},
  booktitle = {Embedded Computer Systems: Architectures, Modeling, and Simulation (SAMOS 2008), Lecture Notes in Computer Science},
  volume    = {5114},
  pages     = {167--176},
  year      = {2008},
  month     = {jul},
  publisher = {Springer},
  doi       = {10.1007/978-3-540-70550-5_19}
}

@article{Sandve2013,
  author  = {Sandve, G. K. and Nekrutenko, A. and Taylor, J. and Hovig, E.},
  title   = {Ten simple rules for reproducible computational research},
  journal = {PLOS Computational Biology},
  volume  = {9},
  number  = {10},
  pages   = {e1003285},
  year    = {2013},
  doi     = {10.1371/journal.pcbi.1003285}
}

@article{StoddenMiguez2014,
  author  = {Stodden, V. and Miguez, S.},
  title   = {Best practices for computational science: Software infrastructure and environments for reproducible and extensible research},
  journal = {Journal of Open Research Software},
  volume  = {2},
  number  = {1},
  pages   = {e21},
  year    = {2014},
  doi     = {10.5334/jors.ay}
}

@article{Jimenez2017,
  author  = {Jim{\'e}nez, R. C. and Kuzak, M. and Alhamdoosh, M. and Barker, M. and Batut, B. and Borg, M. and Capella-Gutierrez, S. and Hong, N. C. and Cook, M. and Corpas, M. and Flannery, M.},
  title   = {Four simple recommendations to encourage best practices in research software},
  journal = {F1000Research},
  volume  = {6},
  pages   = {ELIXIR-876},
  year    = {2017},
  doi     = {10.12688/f1000research.11407.1}
}

@article{Freire2008,
  author  = {Freire, J. and Koop, D. and Santos, E. and Silva, C. T.},
  title   = {Provenance for computational tasks: A survey},
  journal = {Computing in Science \& Engineering},
  volume  = {10},
  number  = {3},
  pages   = {11--21},
  year    = {2008},
  doi     = {10.1109/MCSE.2008.79}
}

@article{Moreau2011,
  author  = {Moreau, L. and Clifford, B. and Freire, J. and Futrelle, J. and Gil, Y. and Groth, P. and Kwasnikowska, N. and Miles, S. and Missier, P. and Myers, J. and Plale, B.},
  title   = {The open provenance model core specification (v1.1)},
  journal = {Future Generation Computer Systems},
  volume  = {27},
  number  = {6},
  pages   = {743--756},
  year    = {2011},
  doi     = {10.1016/j.future.2010.07.005}
}

@inproceedings{Missier2013,
  author    = {Missier, P. and Dey, S. and Belhajjame, K. and Cuevas-Vicentt{\'\i}n, V. and Lud{\"a}scher, B.},
  title     = {{D-PROV}: Extending the {PROV} Provenance Model with {Workflow} Structure},
  booktitle = {5th USENIX Workshop on the Theory and Practice of Provenance (TaPP 13)},
  year      = {2013},
  url       = {https://www.usenix.org/conference/tapp13/technical-sessions/presentation/missier}
}

@article{Boettiger2015,
  author  = {Boettiger, C.},
  title   = {An introduction to Docker for reproducible research},
  journal = {ACM SIGOPS Operating Systems Review},
  volume  = {49},
  number  = {1},
  pages   = {71--79},
  year    = {2015},
  doi     = {10.1145/2723872.2723882}
}

@article{Gruning2018,
  author  = {Gr{\"u}ning, B. and Chilton, J. and K{\"o}ster, J. and Dale, R. and Soranzo, N. and Van Den Beek, M. and Goecks, J. and Backofen, R. and Nekrutenko, A. and Taylor, J.},
  title   = {Practical computational reproducibility in the life sciences},
  journal = {Cell Systems},
  volume  = {6},
  number  = {6},
  pages   = {631--635},
  year    = {2018},
  doi     = {10.1016/j.cels.2018.03.014}
}

@inproceedings{Hannay2009,
  author    = {Hannay, J. E. and MacLeod, C. and Singer, J. and Langtangen, H. P. and Pfahl, D. and Wilson, G.},
  title     = {How do scientists develop and use scientific software?},
  booktitle = {2009 ICSE Workshop on Software Engineering for Computational Science and Engineering},
  pages     = {1--8},
  publisher = {IEEE},
  year      = {2009},
  month     = {may},
  doi       = {10.1109/SECSE.2009.5069155}
}

@misc{Hettrick2014,
  author       = {Hettrick, S. and Antonioletti, M. and Carr, L. and Chue Hong, N. and Crouch, S. and De Roure, D. C. and Emsley, I. and Goble, C. and Hay, A. and Inupakutika, D. and Jackson, M.},
  title        = {UK research software survey 2014},
  howpublished = {Zenodo dataset},
  year         = {2014},
  doi          = {10.5281/zenodo.14809},
  note         = {Dataset}
}

@article{Krafczyk2021,
  author  = {Krafczyk, M. S. and Shi, A. and Bhaskar, A. and Marinov, D. and Stodden, V.},
  title   = {Learning from reproducing computational results: introducing three principles and the Reproduction Package},
  journal = {Philosophical Transactions of the Royal Society A},
  volume  = {379},
  number  = {2197},
  pages   = {20200069},
  year    = {2021},
  doi     = {10.1098/rsta.2020.0069}
}

@article{Ziemann2023,
  author  = {Ziemann, M. and Poulain, P. and Bora, A.},
  title   = {The five pillars of computational reproducibility: bioinformatics and beyond},
  journal = {Briefings in Bioinformatics},
  volume  = {24},
  number  = {6},
  pages   = {bbad375},
  year    = {2023},
  doi     = {10.1093/bib/bbad375}
}

@misc{YildizPeterka2024,
  author       = {Yildiz, O. and Peterka, T.},
  title        = {Do Large Language Models Speak Scientific Workflows?},
  howpublished = {arXiv preprint arXiv:2412.10606},
  year         = {2024},
  doi          = {10.48550/arXiv.2412.10606},
  note         = {arXiv}
}

@misc{AlamRoy2025,
  author       = {Alam, Khairul and Roy, Banani},
  title        = {From Prompt to Pipeline: Large Language Models for Scientific Workflow Development in Bioinformatics},
  howpublished = {arXiv preprint arXiv:2507.20122},
  year         = {2025},
  doi          = {10.48550/arXiv.2507.20122},
  note         = {arXiv}
}

@article{Jiang2025,
  author  = {Jiang, X. and Wang, W. and Tian, S. and Wang, H. and Lookman, T. and Su, Y.},
  title   = {Applications of natural language processing and large language models in materials discovery},
  journal = {npj Computational Materials},
  volume  = {11},
  number  = {1},
  pages   = {79},
  year    = {2025},
  doi     = {10.1038/s41524-025-01554-0}
}

@misc{Yao2022,
  author       = {Yao, S. and Zhao, J. and Yu, D. and Du, N. and Shafran, I. and Narasimhan, K. and Cao, Y.},
  title        = {ReAct: Synergizing Reasoning and Acting in Language Models},
  howpublished = {arXiv preprint arXiv:2210.03629},
  year         = {2022},
  doi          = {10.48550/arXiv.2210.03629}
}

@inproceedings{Wang2024a,
  author    = {Wang, Xingyao and Chen, Yangyi and Yuan, Lifan and Zhang, Yizhe and Li, Yunzhu and Peng, Hao and Ji, Heng},
  title     = {Executable Code Actions Elicit Better {LLM} Agents},
  booktitle = {Proceedings of the 41st International Conference on Machine Learning},
  series    = {Proceedings of Machine Learning Research},
  volume    = {235},
  pages     = {50208--50232},
  year      = {2024},
  url       = {https://proceedings.mlr.press/v235/wang24h.html}
}

@article{Bran2024,
  author  = {Bran, A. M. and Cox, S. and Schilter, O. and Baldassari, C. and White, A. D. and Schwaller, P.},
  title   = {Augmenting large language models with chemistry tools},
  journal = {Nature Machine Intelligence},
  volume  = {6},
  pages   = {525--535},
  year    = {2024},
  doi     = {10.1038/s42256-024-00832-8}
}

@misc{MorishigeKoshihara2025,
  author       = {Morishige, M. and Koshihara, R.},
  title        = {Ensuring Reproducibility in Generative AI Systems for General Use Cases: A Framework for Regression Testing and Open Datasets},
  howpublished = {arXiv preprint arXiv:2505.02854},
  year         = {2025},
  doi          = {10.48550/arXiv.2505.02854}
}

@article{Hosseini2025,
  author  = {Hosseini, M. and Horbach, S. P. and Holmes, K. and Ross-Hellauer, T.},
  title   = {Open Science at the generative AI turn: An exploratory analysis of challenges and opportunities},
  journal = {Quantitative Science Studies},
  volume  = {6},
  pages   = {22--45},
  year    = {2025},
  doi     = {10.1162/qss_a_00337}
}

@misc{Baltes2025,
  author       = {Baltes, S. and Angermeir, F. and Arora, C. and Bar{\'o}n, M. M. and Chen, C. and B{\"o}hme, L. and Calefato, F. and Ernst, N. and Falessi, D. and Fitzgerald, B. and Fucci, D.},
  title        = {Evaluation Guidelines for Empirical Studies in Software Engineering involving Large Language Models},
  howpublished = {arXiv preprint arXiv:2508.15503},
  year         = {2025},
  doi          = {10.48550/arXiv.2508.15503},
  note         = {arXiv}
}

@article{BrucksToubia2025,
  author  = {Brucks, M. and Toubia, O.},
  title   = {Prompt architecture induces methodological artifacts in large language models},
  journal = {PLOS ONE},
  volume  = {20},
  number  = {4},
  pages   = {e0319159},
  year    = {2025},
  doi     = {10.1371/journal.pone.0319159}
}

@misc{Kervadec2023,
  author       = {Kervadec, C. and Franzon, F. and Baroni, M.},
  title        = {Unnatural language processing: How do language models handle machine-generated prompts?},
  howpublished = {arXiv preprint arXiv:2310.15829},
  year         = {2023},
  doi          = {10.48550/arXiv.2310.15829}
}

@article{Ji2023,
  author  = {Ji, Z. and Lee, N. and Frieske, R. and Yu, T. and Su, D. and Xu, Y. and Ishii, E. and Bang, Y. J. and Madotto, A. and Fung, P.},
  title   = {Survey of hallucination in natural language generation},
  journal = {ACM Computing Surveys},
  volume  = {55},
  number  = {12},
  pages   = {1--38},
  year    = {2023},
  doi     = {10.1145/3571730}
}

@misc{Hines2024,
  author       = {Hines, K. and Lopez, G. and Hall, M. and Zarfati, F. and Zunger, Y. and Kiciman, E.},
  title        = {Defending against indirect prompt injection attacks with spotlighting},
  howpublished = {arXiv preprint arXiv:2403.14720},
  year         = {2024},
  doi          = {10.48550/arXiv.2403.14720}
}

@inproceedings{Jia2025,
  author    = {Jia, F. and Wu, T. and Qin, X. and Squicciarini, A.},
  title     = {The task shield: Enforcing task alignment to defend against indirect prompt injection in LLM agents},
  booktitle = {Proceedings of the 63rd Annual Meeting of the Association for Computational Linguistics (Volume 1: Long Papers)},
  pages     = {29680--29697},
  year      = {2025},
  month     = {jul},
  url       = {https://aclanthology.org/2025.acl-long.1435/}
}

@misc{Hu2025,
  author       = {Hu, Y. and Wang, S. and Nie, T. and Zhao, Y. and Wang, H.},
  title        = {Understanding Large Language Model Supply Chain: Structure, Domain, and Vulnerabilities},
  howpublished = {arXiv preprint arXiv:2504.20763},
  year         = {2025},
  doi          = {10.48550/arXiv.2504.20763}
}

@article{Das2025,
  author  = {Das, B. C. and Amini, M. H. and Wu, Y.},
  title   = {Security and privacy challenges of large language models: A survey},
  journal = {ACM Computing Surveys},
  volume  = {57},
  number  = {6},
  pages   = {1--39},
  year    = {2025},
  doi     = {10.1145/3712001}
}

@article{Goecks2010,
  author  = {Goecks, J. and Nekrutenko, A. and Taylor, J. and {Galaxy Team}},
  title   = {Galaxy: a comprehensive approach for supporting accessible, reproducible, and transparent computational research in the life sciences},
  journal = {Genome Biology},
  volume  = {11},
  number  = {8},
  pages   = {R86},
  year    = {2010},
  doi     = {10.1186/gb-2010-11-8-r86}
}

@article{KosterRahmann2012,
  author  = {K{\"o}ster, J. and Rahmann, S.},
  title   = {Snakemake---a scalable bioinformatics workflow engine},
  journal = {Bioinformatics},
  volume  = {28},
  number  = {19},
  pages   = {2520--2522},
  year    = {2012},
  doi     = {10.1093/bioinformatics/bts480}
}

@article{DiTommaso2017,
  author  = {Di Tommaso, P. and Chatzou, M. and Floden, E. W. and Barja, P. P. and Palumbo, E. and Notredame, C.},
  title   = {Nextflow enables reproducible computational workflows},
  journal = {Nature Biotechnology},
  volume  = {35},
  number  = {4},
  pages   = {316--319},
  year    = {2017},
  doi     = {10.1038/nbt.3820}
}

@article{Molder2021,
  author  = {M{\"o}lder, F. and Jablonski, K. P. and Letcher, B. and Hall, M. B. and van Dyken, P. C. and Tomkins-Tinch, C. H. and Sochat, V. and Forster, J. and Vieira, F. G. and Meesters, C. and Lee, S.},
  title   = {Sustainable data analysis with Snakemake},
  journal = {F1000Research},
  volume  = {10},
  pages   = {33},
  year    = {2021},
  doi     = {10.12688/f1000research.29032.3}
}

@article{Kanwal2017,
  author  = {Kanwal, S. and Khan, F. Z. and Lonie, A. and Sinnott, R. O.},
  title   = {Investigating reproducibility and tracking provenance---a genomic workflow case study},
  journal = {BMC Bioinformatics},
  volume  = {18},
  number  = {1},
  pages   = {337},
  year    = {2017},
  doi     = {10.1186/s12859-017-1747-0}
}

@misc{Shin2025,
  author       = {Shin, W. and Souza, R. and Rosendo, D. and Suter, F. and Wang, F. and Balaprakash, P. and da Silva, R. F.},
  title        = {The (R) evolution of Scientific Workflows in the Agentic AI Era: Towards Autonomous Science},
  howpublished = {arXiv preprint arXiv:2509.09915},
  year         = {2025},
  doi          = {10.48550/arXiv.2509.09915}
}

@article{deOliveiraAndrade2025,
  author  = {de Oliveira Andrade, R.},
  title   = {Huge reproducibility project fails to validate dozens of biomedical studies},
  journal = {Nature},
  volume  = {641},
  number  = {8062},
  pages   = {293--294},
  year    = {2025},
  doi     = {10.1038/d41586-025-01266-x}
}

@misc{Anthropic2024MCP,
  author       = {{Anthropic}},
  title        = {Model Context Protocol Specification},
  year         = {2024},
  url          = {https://modelcontextprotocol.io/specification},
  note         = {Accessed: 2025-06-01}
}

@article{Strickland2025,
  author  = {Strickland, J. and Ghisoni, M. and Marshall, H. and Whitehead, T. and Nenchev, B. and Pellegrini, B. and Phillips, C. and Tassenberg, K. and Conduit, G. and Davey, S. and Dorman, S. and Ferguson, D. and Sol, J.},
  title   = {Degrees of uncertainty: conformal deep learning for non-invasive core body temperature prediction in extreme environments},
  journal = {Communications Engineering},
  volume  = {4},
  pages   = {219},
  year    = {2025},
  doi     = {10.1038/s44172-025-00548-6}
}

@article{Strickland2020,
  author  = {Strickland, J. and Nenchev, B. and Perry, S. and Tassenberg, K. and Gill, S. and Panwisawas, C. and Dong, H. and D'Souza, N. and Irwin, S.},
  title   = {On the nature of hexagonality within the solidification structure of single crystal alloys: {M}echanisms and applications},
  journal = {Acta Materialia},
  volume  = {200},
  pages   = {417--431},
  year    = {2020},
  doi     = {10.1016/j.actamat.2020.09.019}
}

@article{Conduit2017,
  author  = {Conduit, B. D. and Jones, N. G. and Stone, H. J. and Conduit, G. J.},
  title   = {Design of a nickel-base superalloy using a neural network},
  journal = {Materials \& Design},
  volume  = {131},
  pages   = {358--365},
  year    = {2017},
  doi     = {10.1016/j.matdes.2017.06.007}
}

@incollection{Strickland2024,
  author    = {Strickland, J. C. J. and Woolston, P. F. D. and Whitehead, T. M.},
  title     = {Adaptive Experimental Design},
  booktitle = {The Digital Transformation of Product Formulation},
  publisher = {CRC Press},
  year      = {2024},
  doi       = {10.1201/9781003385974-15}
}

@article{Ewels2020,
  author  = {Ewels, Philip A. and Peltzer, Alexander and Fillinger, Sven and Patel, Harshil and Alneberg, Johannes and Wilm, Andreas and Garcia, Maxime Ulysse and Di Tommaso, Paolo and Nahnsen, Sven},
  title   = {The nf-core framework for community-curated bioinformatics pipelines},
  journal = {Nature Biotechnology},
  volume  = {38},
  number  = {3},
  pages   = {276--278},
  year    = {2020},
  doi     = {10.1038/s41587-020-0439-x}
}

\end{document}